\documentclass[10pt,twocolumn,letterpaper]{article}

\usepackage[pagenumbers]{cvpr} 

\usepackage{colortbl}
\usepackage{multirow}
\usepackage{svg}

\definecolor{tableblue}{HTML}{DCEDFB}

\definecolor{cvprblue}{rgb}{0.21,0.49,0.74}
\usepackage[pagebackref,breaklinks,colorlinks,allcolors=cvprblue]{hyperref}
\usepackage[accsupp]{axessibility}
\usepackage[T1]{fontenc}
\usepackage{bm}
\usepackage{amsmath}

\def\paperID{} 
\def\confName{CVPR}
\def\confYear{2026}

\title{AGFT: Alignment-Guided Fine-Tuning for Zero-Shot Adversarial Robustness of Vision-Language Models}

\author{Yubo Cui$^1$, Xianchao Guan$^1$, Zijun Xiong$^1$, Zheng Zhang$^{1,2}$\thanks{Corresponding author.}\\
$^1$Harbin Institute of Technology, Shenzhen, China; $^2$Shenzhen Loop Area Institute, Shenzhen, China\\
{\tt\small \{yubo{\_}cui, guanxianchao, 24s051053\}@stu.hit.edu.cn, darrenzz219@gmail.com}
}

\begin{document}
\maketitle

\begin{abstract}
Pre-trained vision-language models (VLMs) exhibit strong zero-shot generalization but remain vulnerable to adversarial perturbations. Existing classification-guided adversarial fine-tuning methods often disrupt pre-trained cross-modal alignment, weakening visual-textual correspondence and degrading zero-shot performance. In this paper, we propose an Alignment-Guided Fine-Tuning (AGFT) framework that enhances zero-shot adversarial robustness while preserving the cross-modal semantic structure. Unlike label-based methods that rely on hard labels and fail to maintain the relative relationships between image and text, AGFT leverages the probabilistic predictions of the original model for text-guided adversarial training, which aligns adversarial visual features with textual embeddings via soft alignment distributions, improving zero-shot adversarial robustness. To address structural discrepancies introduced by fine-tuning, we introduce a distribution consistency calibration mechanism that adjusts the robust model’s output to match a temperature-scaled version of the pre-trained model’s predictions. Extensive experiments across multiple zero-shot benchmarks demonstrate that AGFT outperforms state-of-the-art methods while significantly improving zero-shot adversarial robustness. Code is available at \url{https://github.com/YuboCui/AGFT}.
\end{abstract}

\section{Introduction}
\label{sec:intro}

Pre-trained vision-language models (VLMs) \cite{jia2021scaling,li2022blip,li2025perception}, exemplified by CLIP \cite{radford2021learning}, have emerged as powerful foundation models for multimodal understanding. By leveraging large-scale image-text pairs during pre-training, VLMs learn a shared embedding space that aligns visual representations with textual semantics, enabling strong zero-shot generalization across diverse downstream tasks. However, similar to conventional deep neural networks \cite{szegedy2013intriguing,goodfellow2014explaining,li2025transferable}, VLMs remain vulnerable to adversarial examples \cite{zhao2023evaluating,bhagwatkar2024improving}. Small and imperceptible perturbations can cause severe prediction errors, raising concerns about reliability in practical applications, such as image classification \cite{radford2021learning}, semantic segmentation \cite{zhou2022extract}, and image captioning \cite{chen2022visualgpt}. Thus, improving the adversarial robustness of VLMs is essential for building trustworthy multimodal systems.
\begin{figure}[t]
  \centering

  \begin{minipage}{0.54\linewidth} 
    \begin{subfigure}{\linewidth}
    \centering 
    \includegraphics[width=\linewidth]{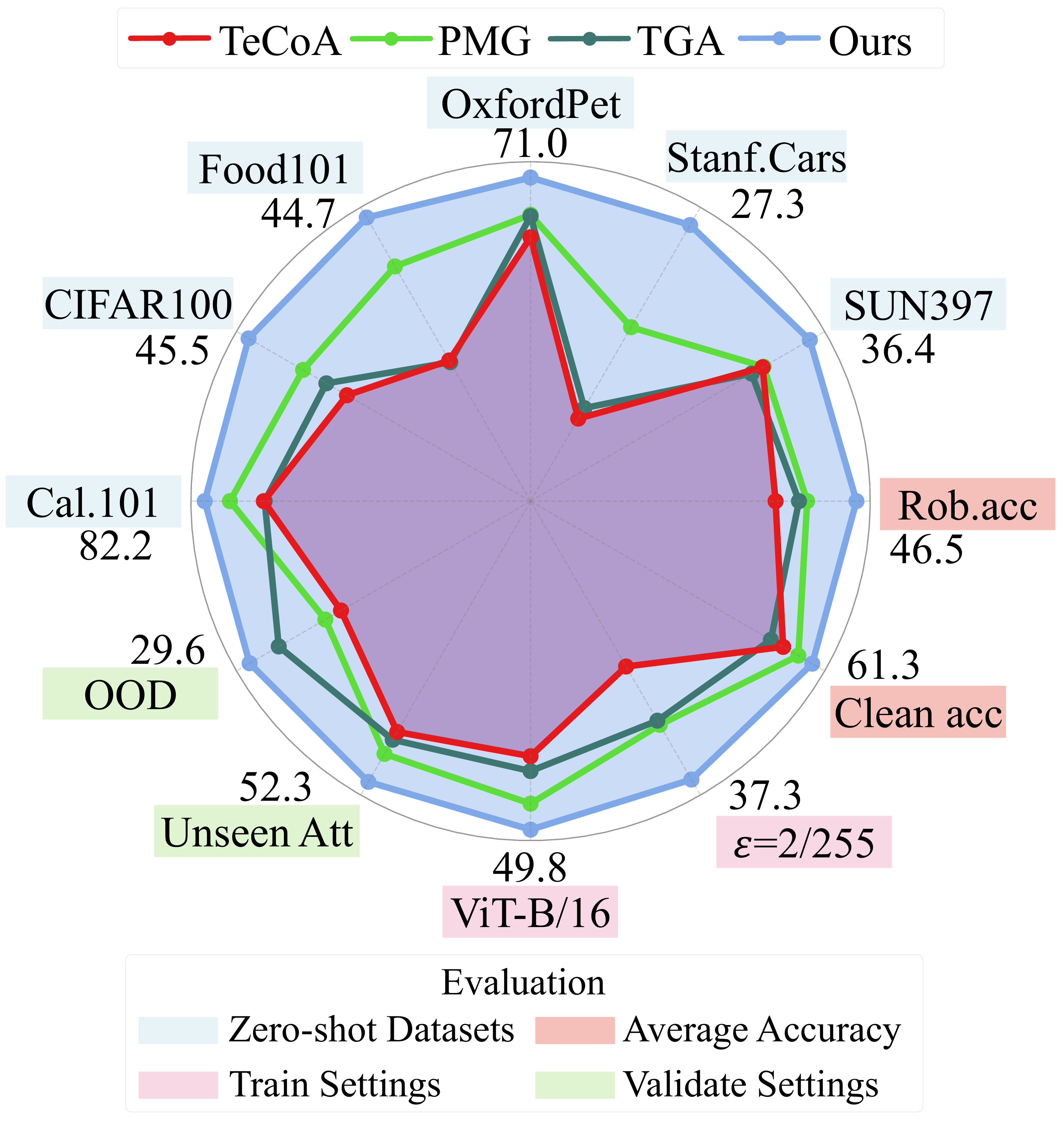}
    \caption{The performance of AGFT}
    \label{fig:short-a}
    \end{subfigure}
  \end{minipage}
  \hfill 
  \begin{minipage}{0.45\linewidth} 
    \centering 
    \begin{subfigure}{\linewidth} 
      \centering
      \includegraphics[width=\linewidth]{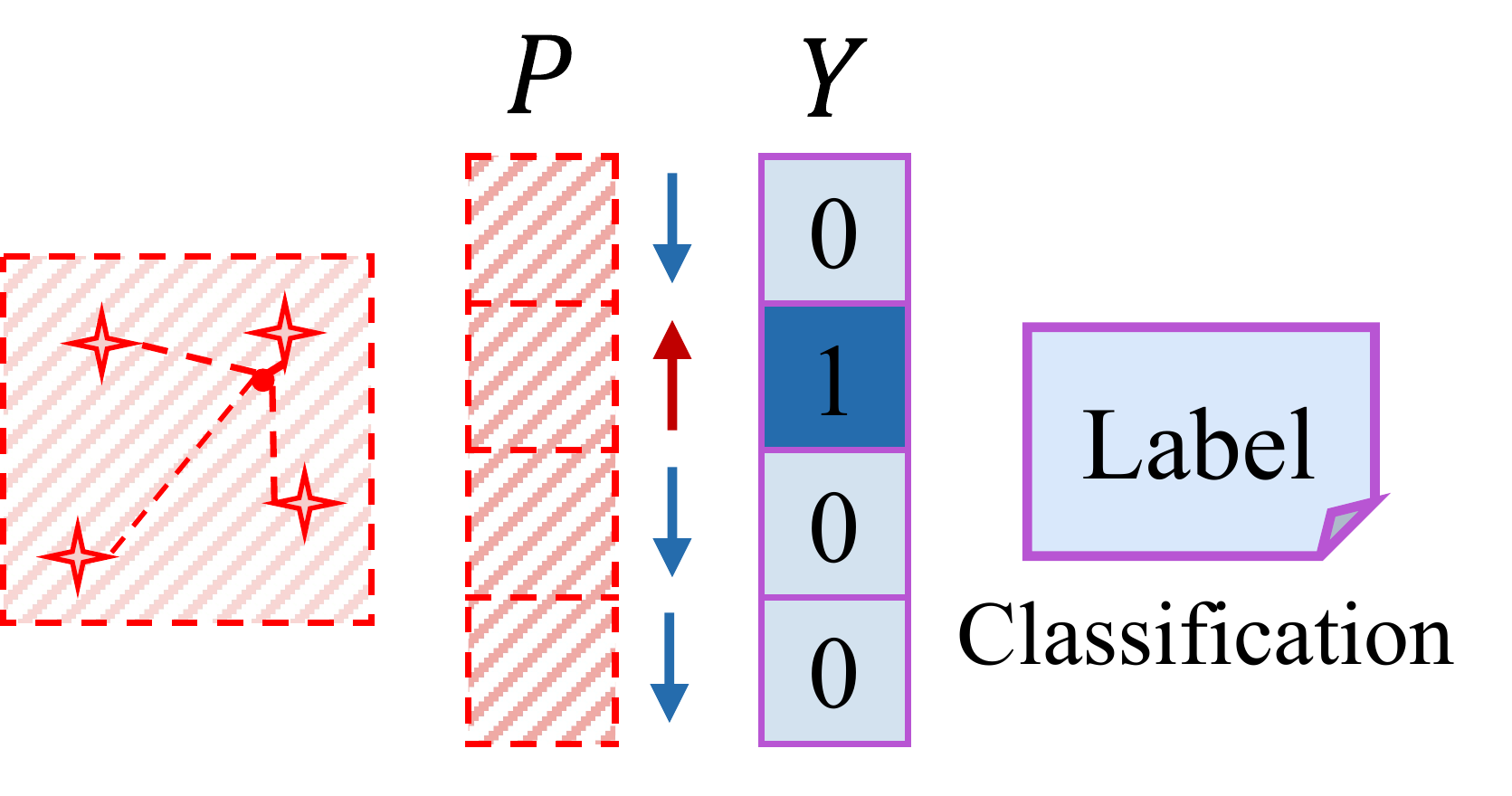}
      \caption{Classification-guided method}
      \label{fig:short-b}
    \end{subfigure}
    \begin{subfigure}{\linewidth} 
      \centering
      \includegraphics[width=\linewidth]{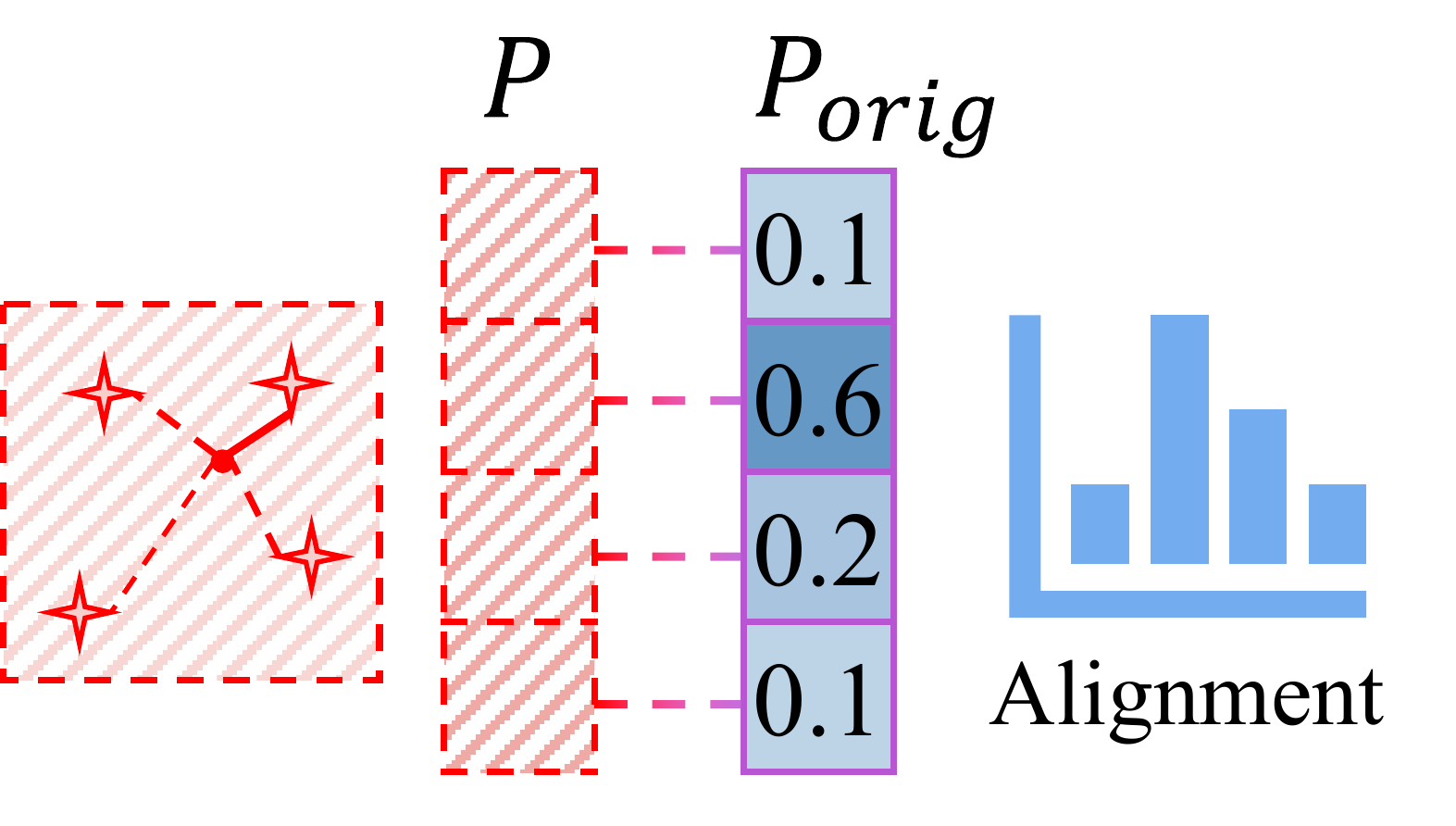}
      \caption{Alignment-guided method}
      \label{fig:short-c}
    \end{subfigure}
    
  \end{minipage}

\caption{The performance of AGFT is shown in \cref{fig:1}(a). Figures \ref{fig:1}(b) and \ref{fig:1}(c) compare classification-guided and alignment-guided adversarial fine-tuning. Unlike classification-guided methods, which rely on label supervision and can disrupt the pre-trained cross-modal alignment, AGFT leverages the probabilistic predictions of the original model to preserve the cross-modal semantic structure, enhancing zero-shot adversarial robustness.}
\label{fig:1} \vspace{-0.5cm}
\end{figure}

Different from the conventional adversarial robustness, achieving Zero-Shot Adversarial Robustness (ZSAR) is particularly challenging \cite{tong2025zero}. VLMs typically recognize by aligning visual features with textual embeddings in a shared space. A robust zero-shot VLM must resist adversarial perturbations while preserving the cross-modal alignment learned during pre-training, which places semantically related image and text representations close together and enables the model to associate unseen categories with their descriptions, generalizing beyond supervised classes \cite{radford2021learning}.

Existing ZSAR approaches \cite{mao2022understanding,dong2025stabilizing} enhance robustness by adversarially fine-tuning the CLIP image encoder under label supervision. These methods treat the pre-trained VLM as a conventional classifier, pushing feature representations toward the target class and suppressing competing classes. While such an approach strengthens class discrimination, it neglects the rich cross-modal similarity correspondence acquired during contrastive pre-training, disrupting the visual-textual alignment \cite{yu2024text}. Consequently, semantic interactions in the shared embedding space are weakened, degrading zero-shot generalization \cite{wang2024pre}. Additionally, adversarial fine-tuning introduces structural biases, leading to a mismatch in prediction distributions and further disrupting the intrinsic cross-modal semantic relationships.

To address these limitations, we propose an Alignment-Guided Fine-Tuning framework to enhance ZSAR while preserving the cross-modal semantic structure of pre-trained VLMs. As shown in Figures \ref{fig:1}(b) and \ref{fig:1}(c), rather than treating a VLM as a conventional classifier, AGFT performs text-guided adversarial training using the original probabilistic predictions as soft supervision. Such an approach enables adversarial visual features to align with textual embeddings according to soft alignment distributions, thereby preserving the pre-trained visual-textual correspondence. To further mitigate structural discrepancies introduced by adversarial fine-tuning, we propose a simple yet effective calibration method that adjusts the robust model's output distribution to match the temperature-scaled predictions of the original model, preserving the similarity structure. Extensive experiments on 15 zero-shot benchmarks across diverse architectures, perturbation budgets, and attack settings demonstrate that AGFT consistently outperforms state-of-the-art methods. Notably, AGFT achieves stronger ZSAR while maintaining the original capabilities of the pre-trained model, as shown in \cref{fig:1}(a). The main contributions of this work are as follows:

\begin{itemize}
\item We propose Alignment-Guided Fine-Tuning (AGFT), a novel framework that enhances zero-shot adversarial robustness in VLMs while maintaining their pre-trained cross-modal semantic structure.
\item We introduce a text-guided adversarial training strategy with temperature-scaled distribution consistency, enhancing zero-shot adversarial robustness by aligning visual features with textual embeddings and mitigating prediction distribution mismatch during fine-tuning.
\item Extensive experiments on various zero-shot benchmarks show that AGFT consistently outperforms state-of-the-art methods in both adversarial robustness and generalization performance.
\end{itemize}

\section{Related Work}
\label{sec:related_work}

\subsection{Adversarial Attack and Defense}
Adversarial attack studies \cite{szegedy2013intriguing} first revealed that imperceptible perturbations to input images can lead to erroneous predictions in deep neural networks \cite{goodfellow2014explaining}, exposing significant security vulnerabilities. Driven by these findings, extensive research on adversarial robustness has emerged, with methods such as DeepFool \cite{moosavi2016deepfool}, the Carlini \& Wagner (C\&W) attack \cite{carlini2017towards}, and AutoAttack \cite{croce2020reliable} systematically evaluating model robustness under adversarial perturbations.

To defend against adversarial attacks, various strategies have been proposed, including detection-based defenses \cite{metzen2017detecting}, purification-based approaches \cite{nie2022diffusion}, and distillation-based methods \cite{papernot2016distillation}. Among these, adversarial training \cite{goodfellow2014explaining,madry2017towards} remains one of the most effective and widely adopted defenses \cite{athalye2018obfuscated}. It is typically framed as a min-max optimization problem, where adversarial examples are generated to maximize the training loss, and the model is optimized to improve robustness. Recent works have further accelerated the training process \cite{shafahi2019adversarial,wong2020fast} and improved the trade-off between robustness and generalization \cite{zhang2019theoretically,wang2019improving}.

\subsection{Adversarial Robustness of VLMs}
Adversarially training VLMs presents unique challenges due to substantial computational overhead and limited generalization in zero-shot settings. Most existing approaches \cite{mao2022understanding,dong2025stabilizing} adopt adversarial fine-tuning, updating only a subset of parameters on top of pre-trained VLMs. TeCoA \cite{mao2022understanding} was the first to explore ZSAR in VLMs, proposing a text-guided contrastive adversarial objective that improves robustness by fine-tuning the CLIP image encoder. PMG-AFT \cite{wang2024pre} builds on TeCoA with an auxiliary branch for supervision, alleviating overfitting during fine-tuning. TGA-ZSR \cite{yu2024text} addresses shifts in text-guided attention caused by adversarial perturbations, introducing attention refinement and model constraints to enhance robustness. GLADIATOR \cite{dong2025stabilizing} further improves robustness by stabilizing the modality gap and reducing gradient norms via effective-rank maximization and feature noise modulation.

Another line of research investigates prompt tuning for adversarial defense in VLMs \cite{li2024one,zhou2025parameter,zhang2024adversarial}. APT \cite{li2024one} shows that adversarial robustness is sensitive to prompt design, learning robust prompts either per category or shared across categories. PEPT \cite{zhou2025parameter} explores word-level prompt search rather than optimizing continuous embeddings, but still depends on robust model weights obtained through adversarial fine-tuning. AdvPT \cite{zhang2024adversarial} learns robust prompts directly from the pre-trained weights, but assumes that attackers cannot access these prompts, limiting its applicability to weaker threat models. Inference-time defenses \cite{sheng2025r,tong2025zero,xing2025clip} avoid adversarial fine-tuning, assuming the defense mechanism is inaccessible to attackers \cite{wang2025tapt}, which restricts their effectiveness against adaptive white-box attacks.

\begin{figure*}[t]
  \centering
  \includegraphics[width=1\linewidth]{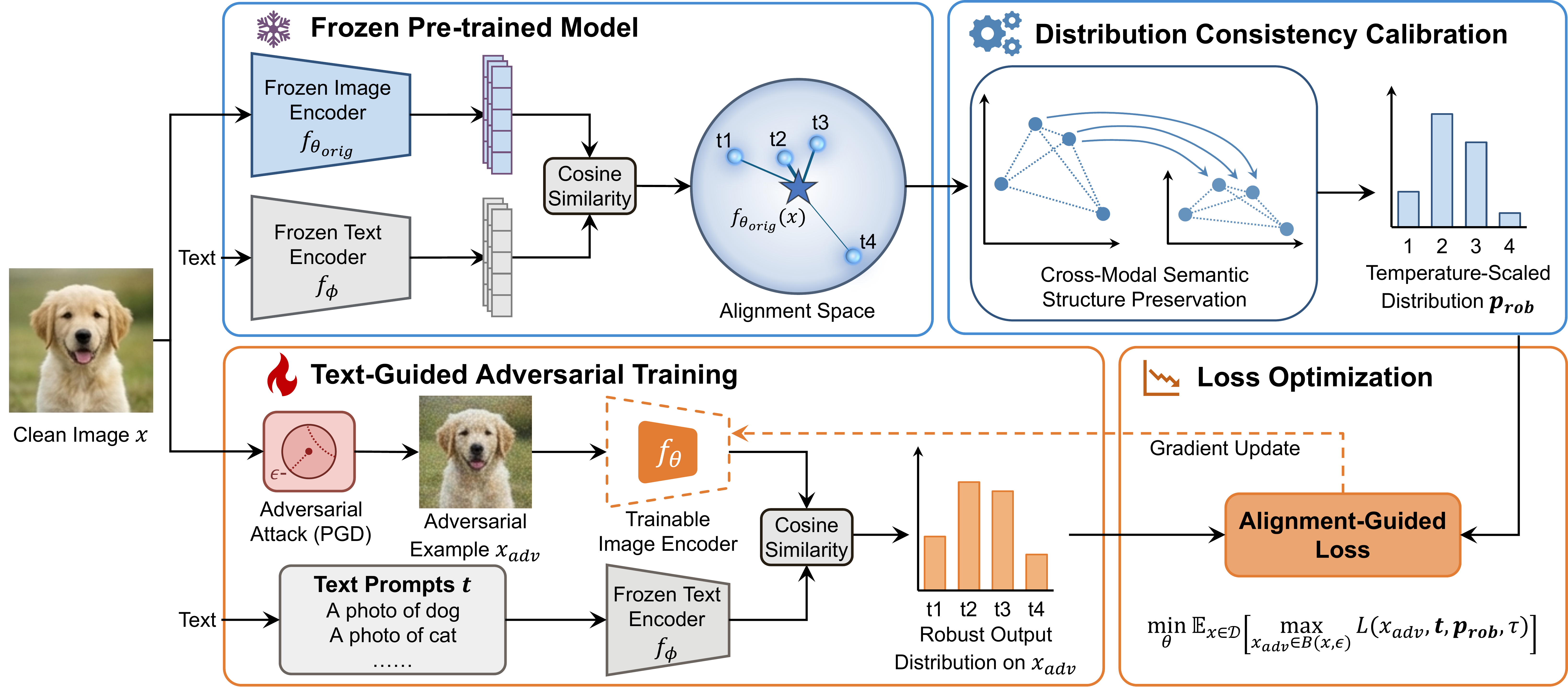}
  \caption{The overall pipeline of AGFT. First, we obtain the probabilistic predictions of the pre-trained model and use the resulting distribution as the target for adversarial fine-tuning to encourage adversarial visual features to align with textual embeddings.
  To mitigate the discrepancies in visual–textual semantic structure, we calibrate the pre-trained output distribution through temperature adjustment, while maintaining the cross-modal similarity structure across images and textual descriptions.}
  \label{fig:2}
\end{figure*}

\section{Methodology}
\label{sec:methodology}


In \cref{sec:preliminaries}, we introduce the conventional CLIP model and its associated notation, followed by a discussion of adversarial attacks, adversarial fine-tuning, and the zero-shot adversarial robustness settings. In \cref{sec:alignment}, we present the two core components of AGFT, \textit{i.e.,} text-guided adversarial training and distribution consistency calibration. The overall pipeline of AGFT is illustrated in \cref{fig:2}.

\subsection{Preliminaries}\label{sec:preliminaries}
\textbf{Notations.} We investigate adversarial fine-tuning for CLIP \cite{radford2021learning}, a foundational vision-language model. CLIP consists of an image encoder $f_{\theta}: \mathcal{X} \to \mathbb{R}^{d}$ and a text encoder $f_{\phi}: \mathcal{T} \to \mathbb{R}^{d}$, where $\theta$ and $\phi$ denote the parameters of two encoders, $\mathcal{X}$ and $\mathcal{T}$ denote the image and text input spaces, respectively, and $d$ is the dimension of the shared embedding space. Given an image $x \in \mathcal{X}$ and a text $t \in \mathcal{T}$, CLIP computes the cosine similarity between the image feature $f_{\theta}(x)$ and the text feature $f_{\phi}(t)$, performing downstream tasks such as zero-shot classification.

CLIP is pre-trained on large-scale image-text pairs using the InfoNCE loss, which pulls matched image-text pairs closer and pushes mismatched pairs apart in the shared embedding space. For a batch of image-text pairs $\{(x^i,t^i)\}_{i=1}^N$, the image-to-text loss is defined as:
\begin{align}
  L_{\text{I-T}} = -\sum_{i=1}^N \log \frac{\exp(\cos(f_{\theta}(x^i), f_{\phi}(t^i))/\tau)}{\sum_{j=1}^N \exp(\cos(f_{\theta}(x^i), f_{\phi}(t^j))/\tau)},
  \label{eq:1}
\end{align}
where $\tau$ is the learnable temperature parameter. The overall loss of CLIP is given by $L_{\text{CLIP}} = (L_{\text{I-T}} + L_{\text{T-I}})/2$, where $L_{\text{T-I}}$ denotes the text-to-image loss.

\noindent\textbf{Adversarial Attacks.} Adversarial attacks aim to mislead the model by adding imperceptible perturbations to the input image, typically constrained by an $\ell_p$-norm. Among existing attack methods, Projected Gradient Descent (PGD) \cite{madry2017towards} is one of the most powerful attack models under white-box settings. PGD iteratively updates the input along the gradient direction that maximizes a given loss function, and then projects the perturbed sample back onto the feasible perturbation set. The generation procedure of PGD is given as follows:
\begin{align}
    \begin{split}
      x_{s+1} &= x_s + \alpha \cdot \text{sign}(\nabla_{x_s} L(x_s,t)), \\
      x_{s+1} &= {\Pi}_{B(x,\epsilon)}(x_{s+1}), \quad s = 0, \dots, S-1,
      \label{eq:2}
    \end{split}
\end{align}
where $x_0=x+\epsilon_0$ with $\epsilon_0 \sim U[-\epsilon, \epsilon]$, $\epsilon$ denotes the perturbation budget, $\alpha$ is the step size, $S$ is the number of iterations, and ${\Pi}_{B(x,\epsilon)}(\cdot)$ denotes the projection operation onto the feasible perturbation set $B(x,\epsilon)=\{x'|\ ||x'-x||_p \leq \epsilon\}$. In this work, we adopt the $\ell_\infty$-norm ball as the constraint.

\noindent\textbf{Zero-Shot Adversarial Robustness.} 
Adversarial training remains  the most effective way to improve the robustness against adversarial attacks. The general objective function is defined as:
\begin{equation}
    \underset{\theta }{\min} \mathbb{E} _{(x,y)\sim \mathcal{D} }\left [ \underset{x+\delta  \in B(x,\epsilon) }{\max}L\left ( f\left ( x+\delta ,\theta  \right )  ,y   \right )  \right ].
    \label{eq:3}
\end{equation}
Such notation follows the standard min-max saddle point formulation introduced by \cite{madry2017towards}. 

In this work, we further study the challenging Zero-Shot Adversarial Robustness (ZSAR) problem, where the target tasks or datasets are outside of training. The goal is to formulate an adversarial training model whose robustness transfers beyond the training distribution and generalizes to these unseen targets without requiring direct access to them. We mainly consider the white-box threat model, like PGD.

\noindent\textbf{Adversarial Fine-tuning.}
Adversarial fine-tuning has been used to improve the ZSAR of VLMs. TeCoA \cite{mao2022understanding} fine-tunes the CLIP image encoder on ImageNet \cite{deng2009imagenet} using class labels to construct text prompts. The training objective is based on cross-entropy optimization:
\begin{align}
  &L(\bm x_{adv}, \bm t, \bm y, \tau) \\
  &= -\mathbb{E}_{i,j} \left[ y^{i,j} \log \frac{\exp\left( \cos\left( f_\theta(x_{adv}^i), f_\phi(t^j) \right)/\tau \right)}{\sum_k \exp\left( \cos\left( f_\theta(x_{adv}^i), f_\phi(t^k) \right)/\tau \right)} \right], \label{eq:4} \nonumber
\end{align}
where $\bm x_{adv}$ denotes the adversarial examples, $\bm t$ denotes the text inputs, and $\bm y$ represents the ground-truth image-text correspondence. Specifically, $y^{i,j} = 1$ indicates positive image-text pair, otherwise, $y^{i,j} = 0$. Herein, the index $k$ runs over all text inputs. $\tau$ is a fixed hyperparameter inherited from pre-training. Accordingly, the adversarial training objective can be formulated as:
\begin{equation}
  \min\mathbb{E}_{\bm x \in \mathcal{D}} \left[ \max_{\bm x_{adv} \in B(\bm x, \epsilon)} L(\bm x_{adv}, \bm t, \bm y, \tau) \right].
  \label{eq:5}
\end{equation}

\subsection{Alignment-Guided Fine-Tuning}
\label{sec:alignment}
Traditional adversarial fine-tuning methods that rely on label supervision often fail to preserve the relative relationships between image and text, disrupting the cross-modal interactions crucial for zero-shot generalization. AGFT improves ZSAR while preserving the cross-modal alignment learned during pre-training. Specifically, text-guided adversarial training maintains the pre-trained feature alignment, and distribution consistency calibration ensures the integrity of the intrinsic cross-modal semantic structure.

\noindent\textbf{Text-guided Adversarial Training.}  
In CLIP, zero-shot recognition is text-guided, with predictions derived from the similarity between image and text embeddings in a shared semantic space. A fundamental prerequisite for enhancing ZSAR is the preservation of original cross-modal correspondence, ensuring perturbed images remain within the semantic neighborhood of their corresponding texts.

Traditional adversarial fine-tuning relies on hard-label supervision, aligning images with a single target text while neglecting the relative similarity across other textual concepts. As a result, the fine-grained semantic correspondence may be severely distorted, particularly under adversarial perturbations. To mitigate such semantic degradation, we replace hard-label supervision with the probabilistic predictions of the pre-trained CLIP model, using the soft distribution for supervision, which helps preserve the relative relationships between images and multiple text embeddings:
\begin{equation}
  p_{{orig}}^{i,j} = \frac{\exp\left(\cos\left(f_{\theta_{{orig}}}(x^i), f_\phi(t^j)\right)/\tau\right)}{\sum_k \exp\left(\cos\left(f_{\theta_{{orig}}}(x^i), f_\phi(t^k)\right)/\tau\right)}.
  \label{eq:6}
\end{equation}
Leveraging the original probabilistic predictions, we align the adversarial visual features with text embeddings via soft distributions, maintaining the intrinsic semantic correspondence while improving ZSAR.

\begin{table*}[t]
\caption{Zero-shot \textbf{robust} accuracy (\%).
Adversarial examples are generated by PGD-20 attack with
the perturbation budget $\epsilon = 1/255$.
Models are adversarially fine-tuned on ImageNet and evaluated across 15 datasets.
Results of GLADIATOR are from the original paper.}
\vspace{-9pt} 
\centering
\resizebox{\linewidth}{!}{%
\renewcommand{\arraystretch}{0.9} 
\begin{tabular}{c c c c c c c c c c c c c c c c c}
\toprule
\textbf{Method} & \rotatebox{90}{Caltech101} & \rotatebox{90}{Caltech256} & \rotatebox{90}{CIFAR10} & \rotatebox{90}{CIFAR100} & \rotatebox{90}{DTD} & \rotatebox{90}{EuroSAT} & \rotatebox{90}{FGVC} & \rotatebox{90}{Flower102} & \rotatebox{90}{Food101} & \rotatebox{90}{ImageNet} & \rotatebox{90}{OxfordPet} & \rotatebox{90}{PCAM} & \rotatebox{90}{Stanf.Cars} & \rotatebox{90}{STL10} & \rotatebox{90}{SUN397} & {\cellcolor{tableblue}\textbf{Average}} \\
\midrule
CLIP & 21.27 & 11.69 & 10.31 & 4.69 & 2.55 & 0.02 & 0.00 & 1.00 & 4.06 & 1.13 & 2.37 & 0.13 & 0.21 & 33.10 & 1.08 & \cellcolor{tableblue}6.24 \\
\midrule
TeCoA & 71.83 & 61.23 & 59.85 & 34.47 & 23.56 & 15.06 & 4.85 & 30.60 & 29.01 & 41.29 & 61.90 & 14.42 & 14.06 & 83.33 & 32.12 & \cellcolor{tableblue}38.51 \\
PMG-AFT & 77.78 & 63.15 & 65.93 & 39.41 & 26.70 & 15.05 & 5.93 & 33.36 & 39.37 & 35.05 & 65.34 & 18.99 & 20.33 & 86.31 & 32.21 & \cellcolor{tableblue}41.66 \\
TGA-ZSR & 71.67 & 64.06 & 63.30 & 36.79 & 22.07 & 13.95 & 4.85 & 27.26 & 28.81 & 52.80 & 65.12 & 30.95 & 14.78 & 85.01 & 31.12 & \cellcolor{tableblue}40.84 \\
GLADIATOR & 73.34 & 63.50 & 67.89 & 40.82 & 24.75 & 14.92 & 6.23 & 34.58 & 34.92 & 44.53 & 68.10 & 39.29 & 20.65 & 86.53 & 31.86 & \cellcolor{tableblue}43.46 \\
\cellcolor{tableblue}\textbf{AGFT (Ours)} & \cellcolor{tableblue}{82.23} & \cellcolor{tableblue}{67.95} & \cellcolor{tableblue}{71.72} & \cellcolor{tableblue}{45.55} & \cellcolor{tableblue}{24.14} & \cellcolor{tableblue}{16.25} & \cellcolor{tableblue}{7.94} & \cellcolor{tableblue}{35.95} & \cellcolor{tableblue}{44.76} & \cellcolor{tableblue}{44.95} & \cellcolor{tableblue}{71.03} & \cellcolor{tableblue}{33.89} & \cellcolor{tableblue}{27.34} & \cellcolor{tableblue}{88.52} & \cellcolor{tableblue}{36.39} & \cellcolor{tableblue}\textbf{46.57} \\
\bottomrule
\end{tabular}%
}
\label{tab:1}
\end{table*}

\begin{table*}[t]
\captionsetup{justification=justified, singlelinecheck=false}
\caption{Zero-shot \textbf{clean} accuracy (\%).
Clean images from 15 datasets are evaluated on adversarially fine-tuned CLIP models.}
\vspace{-9pt} 
\centering
\resizebox{\linewidth}{!}{%
\renewcommand{\arraystretch}{0.9} 
\begin{tabular}{c c c c c c c c c c c c c c c c c}
\toprule
\textbf{Method} & \rotatebox{90}{Caltech101} & \rotatebox{90}{Caltech256} & \rotatebox{90}{CIFAR10} & \rotatebox{90}{CIFAR100} & \rotatebox{90}{DTD} & \rotatebox{90}{EuroSAT} & \rotatebox{90}{FGVC} & \rotatebox{90}{Flower102} & \rotatebox{90}{Food101} & \rotatebox{90}{ImageNet} & \rotatebox{90}{OxfordPet} & \rotatebox{90}{PCAM} & \rotatebox{90}{Stanf.Cars} & \rotatebox{90}{STL10} & \rotatebox{90}{SUN397} & {\cellcolor{tableblue}\textbf{Average}} \\
\midrule
CLIP & 92.03 & 82.81 & 87.19 & 61.15 & 41.17 & 45.40 & 19.25 & 65.34 & 83.61 & 59.53 & 87.32 & 58.26 & 54.10 & 96.89 & 59.01 & \cellcolor{tableblue}66.20 \\
\midrule
TeCoA & 85.30 & 77.34 & 78.75 & 50.02 & 34.68 & {27.22} & 12.62 & 51.31 & 56.24 & 63.50 & 82.01 & 53.44 & 33.88 & 93.23 & 54.45 & \cellcolor{tableblue}56.93 \\
PMG-AFT & 90.25 & 78.33 & 83.24 & 57.26 & 35.42 & 25.93 & 13.97 & 52.79 & 70.03 & 53.57 & 82.58 & 54.22 & 42.21 & 94.73 & 53.96 & \cellcolor{tableblue}59.23 \\
TGA-ZSR & 81.70 & 77.45 & 81.18 & 53.45 & 31.70 & 24.81 & 9.98 & 44.19 & 51.31 & {70.96} & 79.91 & 46.48 & 29.34 & 93.52 & 50.10 & \cellcolor{tableblue}55.07 \\
GLADIATOR & 85.42 & 79.44 & {86.94} & 60.86 & {35.59} & 26.52 & 15.66 & {56.77} & 67.32 & 63.83 & 84.25 & 49.37 & 43.75 & 95.71 & 53.61 & \cellcolor{tableblue}60.34 \\
\cellcolor{tableblue}\textbf{AGFT (Ours)} & \cellcolor{tableblue}{91.05} & \cellcolor{tableblue}{80.26} & \cellcolor{tableblue}86.72 & \cellcolor{tableblue}{62.38} & \cellcolor{tableblue}33.40 & \cellcolor{tableblue}25.06 & \cellcolor{tableblue}{16.49} & \cellcolor{tableblue}54.89 & \cellcolor{tableblue}{72.14} & \cellcolor{tableblue}60.11 & \cellcolor{tableblue}{85.58} & \cellcolor{tableblue}{55.52} & \cellcolor{tableblue}{45.69} & \cellcolor{tableblue}{95.72} & \cellcolor{tableblue}{55.19} & \cellcolor{tableblue}\textbf{61.35} \\
\bottomrule
\end{tabular}%
}
\vspace{-6pt} 
\label{tab:2}
\end{table*}

\noindent\textbf{Distribution Consistency Calibration.}  
Although text-guided adversarial fine-tuning already improves over label supervision by leveraging $\bm p_{\text{orig}}$, directly matching the pre-trained distribution remains suboptimal.
The reason is that $\bm p_{\text{orig}}$ entangles two factors: the relative image-text similarity structure and the confidence scale inherited from the clean pre-trained model.
The former captures how an image distributes its semantic similarity across all candidate text prompts, beyond the ground-truth label, while the latter is reflected in the magnitudes of the prediction logits.
Directly using $\bm p_{\text{orig}}$ as the target in adversarial training therefore forces the robust model to inherit not only the desired semantic relations but also the pre-trained confidence scale, and the latter may be mismatched to the evolving robust feature space, thereby interfering with the preservation of the original cross-modal structure.

To mitigate such structural distortions, we introduce distribution consistency calibration to align the robust model to match the cross-modal semantic structure of the original model. Specifically, it should maintain the relative semantic relationships between the image and multiple texts, i.e.,
\begin{equation}
  \frac
  {S_{rob}(i,j)}
  {\sum_k S_{rob}(i,k)}
  \leftrightarrow 
  \frac
  {S_{orig}(i,j)}
  {\sum_k S_{orig}(i,k)},
  \label{eq:7}
\end{equation}
where $S_{orig}(i,j)$ and $S_{rob}(i,j)$ are the similarities between the $i$-th image and $j$-th text in the original pre-trained model and the robust model, respectively.

Specifically, to maintain distribution consistency, we adjust the confidence scale of the robust model’s predictions by introducing a scaled temperature parameter to the logits of the original pre-trained model. The temperature adjustment reduces overconfident predictions, ensuring that the model focuses on stable semantic features instead of noisy or adversarial perturbations. The resulting temperature-scaled target distribution preserves cross-modal semantic consistency while mitigating excessive reliance on high-confidence logits. Consequently, the robust model is guided by a target distribution better matched to the robust feature space, while preserving the global cross-modal semantic structure of the original model. We define the distribution consistency calibration module as:
\begin{equation}
\begin{split}
  p_{{rob}}^{i,j} &= \frac{\exp\left(\cos\left(f_{\theta_{{orig}}}(x^i), f_\phi(t^j)\right)/(\tau/\gamma)\right)}{\sum_k \exp\left(\cos\left(f_{\theta_{{orig}}}(x^i), f_\phi(t^k)\right)/(\tau/\gamma)\right)},
  \label{eq:8}
\end{split}
\end{equation}
where $\gamma \in (0,1]$ denotes the temperature scaling ratio, and the calibrated temperature is $\tau / \gamma$. Here, $\gamma$ is used only to calibrate the pre-trained target distribution, while both calibration and adversarial fine-tuning share $\tau$. In practice, $\tau$ is tuned to control the distribution sharpness, facilitating robust feature learning during fine-tuning.

\noindent\textbf{The Objective Function of AGFT.} Following the overall scheme of text-guided adversarial training,
we replace $ p_{{orig}}^{i,j}$ with the calibrated target distribution $ p_{{rob}}^{i,j}$ in the training objective, which is formulated as:

\begin{align}
  &L(\bm x_{adv}, \bm t, \bm p_{{rob}}, \tau) =  \\
  &-\mathbb{E}_{i,j} \left[ p_{{rob}}^{i,j} \log \frac{\exp\left( \cos\left( f_\theta(x_{adv}^i), f_\phi(t^j) \right)/\tau \right)}{\sum_k \exp\left( \cos\left( f_\theta(x_{adv}^i), f_\phi(t^k) \right)/\tau \right)} \right], \label{eq:9} \nonumber
\end{align}
Therefore, we define the final alignment-guided adversarial fine-tuning objective function of AGFT as a min-max optimization problem, \textit{i.e.,}
\begin{equation}
  \min \mathbb{E}_{\bm x \in \mathcal{D}} \left[ \max_{\bm x_{adv} \in B(\bm x, \epsilon)} L(\bm x_{adv}, \bm t, \bm p_{{rob}}, \tau) \right].
  \label{eq:10}
\end{equation}


\section{Experiments}
\label{sec:Experiments}

\subsection{Experimental Setup}
\label{sec:experimental}
\textbf{Datasets.}
We adversarially fine-tune CLIP on the ImageNet \cite{deng2009imagenet} training set, following previous works \cite{mao2022understanding, dong2025stabilizing}. We evaluate both clean and adversarial performance on 15 benchmarks, which include the ImageNet validation set and 14 additional zero-shot transfer datasets across five task families:
generic classification on
Caltech101 \cite{fei2004learning},
Caltech256 \cite{griffin2007caltech},
CIFAR10 \cite{krizhevsky2009learning},
CIFAR100 \cite{krizhevsky2009learning},
and STL10 \cite{coates2011analysis};
fine-grained classification on
FGVCAircraft \cite{maji2013fine},
Flowers102 \cite{nilsback2008automated},
Food101 \cite{bossard2014food},
OxfordPets \cite{parkhi2012cats},
and StanfordCars \cite{krause20133d};
scene recognition on
SUN397 \cite{xiao2010sun};
texture recognition on
DTD \cite{cimpoi2014describing};
domain-specialized classification on
EuroSAT \cite{helber2019eurosat}
and PCAM \cite{veeling2018rotation}.
For out-of-distribution evaluation, we also consider two variants of ImageNet: ImageNet-R \cite{hendrycks2021many}, which features altered textures and local statistics, and ImageNet-S \cite{wang2019learning}, containing sketch-style images.

\noindent\textbf{Baselines.}  
We compare AGFT with four representative adversarial fine-tuning baselines: TeCoA \cite{mao2022understanding}, PMG-AFT \cite{wang2024pre}, TGA-ZSR \cite{yu2024text}, and GLADIATOR \cite{dong2025stabilizing}. For each method, we report both clean accuracy and robust accuracy across multiple training and evaluation settings, allowing for a comprehensive assessment of AGFT’s effectiveness. Detailed implementation settings for all baselines are provided in the Appendix (\cref{sec:baselines}).

\noindent\textbf{Implementation Details.}  
We adopt the CLIP model with a ViT-B/32 \cite{dosovitskiy2020image} image encoder. We fine-tune all parameters of the image encoder while keeping the text encoder frozen. Training is conducted for 10 epochs using SGD with an initial learning rate of $4\times10^{-4}$, momentum of 0.9, cosine learning-rate decay, and a batch size of 256. During adversarial fine-tuning, adversarial examples are generated using a 2-step PGD with perturbation budgets $\epsilon \in \{1/255, 2/255, 4/255\}$, where the step size is set to $\alpha=\epsilon$. For robustness evaluation, we employ a 20-step PGD with a step size of $\alpha=1/255$. The hyperparameters are set as $\gamma = 0.4$ and $\tau = 1/180$. All experiments are conducted on two NVIDIA GeForce RTX 3090 GPUs.

\noindent\textbf{Attack Configuration.}  
Beyond standard PGD  \cite{madry2017towards}, we evaluate robustness using  two stronger white-box attacks: the Carlini \& Wagner (C\&W) attack \cite{carlini2017towards} and AutoAttack (AA) \cite{croce2020reliable}. To assess generalizability against unseen threats, we further include several attack methods: MI-FGSM \cite{dong2018boosting}, DI\textsuperscript{2}-FGSM \cite{xie2019improving}, TI-FGSM \cite{dong2019evading}, NI-FGSM \cite{lin2019nesterov}, PI-FGSM \cite{gao2020patch}, and PI-FGSM++ \cite{gao2020patch2}. Detailed attack settings are provided in the Appendix (\cref{sec:attack}).

\begin{table}[t]
\caption{Average zero-shot robust accuracy (\%) under PGD-20 attacks with varying perturbation budgets  during \textbf{inference}.
}
\vspace{-9pt} 
\centering
\footnotesize 
\renewcommand{\arraystretch}{0.9}
\begin{tabular}{c c c c c c}
\toprule
\multirow{2}{*}[-0.6ex]{Method} & \multicolumn{5}{c}{Robust Accuracy} \\
\cmidrule{2-6}
& 1/255 & 2/255 & 3/255 & 4/255 & Avg. \\
\midrule
TeCoA       & 38.51 & 22.30 & 11.89 & 6.75  & 19.86 \\
PMG-AFT      & 41.66 & 25.15 & 13.63 & 7.19  & 21.90 \\
TGA-ZSR       & 40.84  & 26.61 & 16.15 & 10.30 & 23.47 \\
GLADIATOR       & 43.46 & 26.41 & 15.29 & 8.59 & 23.43 \\
\rowcolor{tableblue}
\textbf{AGFT (Ours)} & {46.57} & {29.80} & {16.61} & {8.88} & \textbf{25.46} \\
\bottomrule
\end{tabular}
\vspace{-6pt} 
\label{tab:3}
\end{table}

\begin{table}[t]
\caption{Average accuracy (\%) under strong attacks with different perturbation budgets used during both \textbf{fine-tuning and inference}.}
\vspace{-9pt} 
\renewcommand{\arraystretch}{0.9}
\centering
\footnotesize 

\begin{tabular}{c c c c c c}
\toprule
{Budget $\epsilon$} & {Method} & {Clean} & {PGD} & {C\&W} & {AA} \\
\midrule

\multirow{5}{*}{{1/255}} 
& TeCoA     & 56.93 & 38.51 & 37.57 & 37.01 \\
& PMG-AFT    & 59.23 & 41.66 & 40.55 & 40.09 \\
& TGA-ZSR      & 55.07 & 40.84 & 40.26 & 39.90 \\
& GLADIATOR     & 60.34 & 43.46 & 42.50 & 41.91 \\
& \cellcolor{tableblue}\textbf{AGFT (Ours)} & \cellcolor{tableblue}\textbf{61.35} & \cellcolor{tableblue}\textbf{46.57} & \cellcolor{tableblue}\textbf{43.06} & \cellcolor{tableblue}\textbf{42.81} \\
\midrule

\multirow{5}{*}{{2/255}} 
& TeCoA     & 49.69 & 26.79 & 25.92 & 25.14 \\
& PMG-AFT    & 54.69 & 32.22 & 31.32 & 30.61 \\
& TGA-ZSR      & 50.09 & 31.83 & 31.30 & 30.74 \\
& GLADIATOR     & 51.24 & 30.56 & 29.37 & 28.49 \\
& \cellcolor{tableblue}\textbf{AGFT (Ours)} & \cellcolor{tableblue}\textbf{56.86} & \cellcolor{tableblue}\textbf{37.31} & \cellcolor{tableblue}\textbf{33.79} & \cellcolor{tableblue}\textbf{33.34} \\
\midrule

\multirow{5}{*}{{4/255}} 
& TeCoA     & 38.20 & 14.84 & 13.97 & 13.22 \\
& PMG-AFT    & 34.58 & 6.69 & 6.20 & 5.18 \\
& TGA-ZSR      & 41.73 & 21.14 & 20.65 & 19.82 \\
& GLADIATOR     & 38.10 & 18.29 & 16.24 & 13.76 \\
& \cellcolor{tableblue}\textbf{AGFT (Ours)} & \cellcolor{tableblue}\textbf{49.47} & \cellcolor{tableblue}\textbf{24.98} & \cellcolor{tableblue}\textbf{21.89} & \cellcolor{tableblue}\textbf{21.25} \\
\bottomrule
\end{tabular}
\vspace{-6pt} 
\label{tab:4}
\end{table}

\subsection{Main Results}
\label{sec:main}
\noindent\textbf{Performance across 15 Datasets.}  
Following the standard protocol for ZSAR \cite{mao2022understanding, wang2024pre}, we adversarially fine-tune the model with a perturbation budget of $\epsilon = 1/255$ in the main setting. The robust and clean accuracy across 15 datasets are reported in \cref{tab:1} and \cref{tab:2}, respectively.

While the pre-trained CLIP model demonstrates strong zero-shot generalization on clean images, it remains highly vulnerable to adversarial perturbations. AGFT achieves the best average performance on both metrics, surpassing the strongest baseline by 3.1\% in robust accuracy and 1.0\% in clean accuracy. Compared to the classification-guided TeCoA, AGFT improves robust and clean accuracy by 8.1\% and 4.4\%, respectively. These results highlight that AGFT significantly enhances ZSAR while better preserving the clean-data generalization of the pre-trained model.

\noindent\textbf{Robustness across Varying Perturbation Budgets.}  
In addition to the standard evaluation setting, we further test the model fine-tuned with $\epsilon = 1/255$ under varying perturbation budgets at inference time. As shown in \cref{tab:3}, AGFT achieves the best average robust accuracy across different perturbation budgets, demonstrating stronger robustness generalization. While TGA-ZSR slightly outperforms AGFT at $\epsilon = 4/255$, AGFT consistently performs better at smaller perturbation budgets and remains superior on average. Overall, these results suggest that AGFT generalizes more effectively across varying attack strengths.

\begin{table}[t]
\caption{Average performance (\%) of \textbf{different CLIP architectures}
with the perturbation budget $\epsilon = 1/255$.}
\vspace{-9pt} 
\renewcommand{\arraystretch}{0.9}
\centering
\footnotesize 

\begin{tabular}{c c c c c c}
\toprule
{Architecture} & {Method} & {Clean} & {PGD} & {C\&W} & {AA} \\
\midrule

\multirow{4}{*}{{ViT-B/32}} 
& TeCoA     & 56.93 & 38.51 & 37.57 & 37.01 \\
& PMG-AFT    & 59.23 & 41.66 & 40.55 & 40.09 \\
& TGA-ZSR      & 55.07 & 40.84 & 40.26 & 39.90 \\
& \cellcolor{tableblue}\textbf{AGFT (Ours)} & \cellcolor{tableblue}\textbf{61.35} & \cellcolor{tableblue}\textbf{46.57} & \cellcolor{tableblue}\textbf{43.06} & \cellcolor{tableblue}\textbf{42.81} \\
\midrule

\multirow{4}{*}{{ViT-B/16}} 
& TeCoA     & 60.12 & 42.05 & 41.15 & 40.50 \\
& PMG-AFT    & 61.49 & 47.08 & 46.08 & 45.60 \\
& TGA-ZSR      & 57.39 & 43.63 & 43.12 & 42.76 \\
& \cellcolor{tableblue}\textbf{AGFT (Ours)} & \cellcolor{tableblue}\textbf{63.03} & \cellcolor{tableblue}\textbf{49.84} & \cellcolor{tableblue}\textbf{46.68} & \cellcolor{tableblue}\textbf{46.38} \\
\midrule

\multirow{4}{*}{{RN50$\times$4}} 
& TeCoA     & \textbf{52.04} & 31.00 & 30.30 & 28.88 \\
& PMG-AFT    & 49.23 & 30.26 & 28.71 & 28.18 \\
& TGA-ZSR      & - & - & - & - \\
& \cellcolor{tableblue}\textbf{AGFT (Ours)} & \cellcolor{tableblue}{51.23} & \cellcolor{tableblue}\textbf{34.69} & \cellcolor{tableblue}\textbf{30.92} & \cellcolor{tableblue}\textbf{30.56} \\
\bottomrule
\end{tabular}
\vspace{-6pt} 
\label{tab:6}
\end{table}

\noindent\textbf{Fine-tuning with Larger Perturbation Budgets.}  
To evaluate different defense regimes, we also adversarially fine-tune models with larger perturbation budgets. The results in \cref{tab:4} show that AGFT consistently outperforms all evaluated  baselines in both clean and robust accuracy across various training radii. In particular, under $\epsilon = 4/255$, AGFT improves clean accuracy by 7.7\% and robust accuracy by 3.8\% over the best baseline. Such consistent performance confirms that AGFT remains highly  effective even under stronger adversarial training settings.

\noindent\textbf{Robustness against Stronger Attacks.}  
In addition to the standard  PGD-20 attack, we further evaluate adversarial  robustness under two stronger white-box attacks: the Carlini \& Wagner (C\&W) attack \cite{carlini2017towards} and AutoAttack (AA) \cite{croce2020reliable}. AGFT consistently achieves the best robust accuracy under both attacks across various perturbation budgets. Such broad generalization confirms that the improvements of AGFT are not limited to PGD evaluations and generalize well to stronger attack scenarios. Detailed per-dataset results are provided in the Appendix (\cref{sec:additional}).

\noindent\textbf{Robustness against Diverse Unseen Attacks.}  
To further evaluate robustness generalization beyond the specific attack configuration used during training, we consider a diverse set of attack methods, including MI-FGSM \cite{dong2018boosting}, DI\textsuperscript{2}-FGSM \cite{xie2019improving}, TI-FGSM \cite{dong2019evading}, NI-FGSM \cite{lin2019nesterov}, PI-FGSM \cite{gao2020patch}, and PI-FGSM++ \cite{gao2020patch2}. In addition to untargeted attacks, we also evaluate targeted attacks with two target selection strategies: random target selection and least-likely target selection. As shown in \cref{tab:5}, AGFT achieves the best average robust accuracy across all 15 attack settings, outperforming the strongest baseline by 3.6\%. These results confirm that AGFT provides stronger robustness generalization across diverse unseen attack types.

\begin{table*}[t]
\caption{Zero-shot robust accuracy (\%) under diverse \textbf{unseen untargeted and targeted attacks}
with the perturbation budget $\epsilon = 1/255$.
Targeted attacks with random target selection are denoted by
ATT\textsubscript{T}, and those with least-likely target selection are denoted by
ATT\textsubscript{TL}.}
\vspace{-9pt} 
\centering
\resizebox{\linewidth}{!}{%
\renewcommand{\arraystretch}{0.9} 
\begin{tabular}{c c c c c c c c c c c c c c c c c}
\toprule
\textbf{Method} & PGD & MI & DI\textsuperscript{2} & TI & NI & PI & PI++ & C\&W & AA & PGD\textsubscript{T} & PGD\textsubscript{TL} & MI\textsubscript{T} & DI\textsuperscript{2}\textsubscript{T} & TI\textsubscript{T} & NI\textsubscript{T} & {\cellcolor{tableblue}\textbf{Average}} \\
\midrule
TeCoA & 38.51 & 38.53 & 41.93 & 47.13 & 49.45 & 38.80 & 40.06 & 37.57 & 37.01 & 51.93 & 52.49 & 51.90 & 53.02 & 54.55 & 55.15 & \cellcolor{tableblue}45.86 \\
PMG-AFT & 41.66 & 41.69 & 44.82 & 49.21 & 52.08 & 41.93 & 42.84 & 40.55 & 40.09 & 54.96 & 55.42 & 54.96 & 55.83 & 57.02 & 57.67 & \cellcolor{tableblue}48.71 \\
TGA-ZSR & 40.84 & 40.88 & 43.43 & 47.18 & 48.92 & 41.07 & 41.97 & 40.26 & 39.90 & 52.36 & 52.92 & 52.33 & 53.00 & 53.85 & 54.14 & \cellcolor{tableblue}46.87 \\
\cellcolor{tableblue}\textbf{AGFT (Ours)} & \cellcolor{tableblue}\textbf{46.57} & \cellcolor{tableblue}\textbf{46.56} & \cellcolor{tableblue}\textbf{48.99} & \cellcolor{tableblue}\textbf{53.26} & \cellcolor{tableblue}\textbf{55.68} & \cellcolor{tableblue}\textbf{46.73} & \cellcolor{tableblue}\textbf{47.93} & \cellcolor{tableblue}\textbf{43.06} & \cellcolor{tableblue}\textbf{42.81} & \cellcolor{tableblue}\textbf{57.99} & \cellcolor{tableblue}\textbf{58.41} & \cellcolor{tableblue}\textbf{58.01} & \cellcolor{tableblue}\textbf{58.87} & \cellcolor{tableblue}\textbf{59.92} & \cellcolor{tableblue}\textbf{60.42} & \cellcolor{tableblue}\textbf{52.34} \\
\bottomrule
\end{tabular}%
}
\vspace{-6pt} 
\label{tab:5}
\end{table*}

\noindent\textbf{Fine-tuning across Different Architectures.}  
To evaluate the architectural generality of AGFT, we test it on two additional CLIP backbones: ViT-B/16 \cite{dosovitskiy2020image} and RN50$\times$4 \cite{he2016deep}. As shown in \cref{tab:6}, AGFT consistently delivers strong performance across architectures, demonstrating its effectiveness is not limited to a specific backbone design. For the ViT-based models, AGFT achieves the best clean and robust accuracy across all evaluated attacks. For RN50$\times$4, while TeCoA slightly outperforms AGFT in clean accuracy, AGFT achieves the highest robustness under PGD, C\&W attack, and AutoAttack, resulting in better overall robustness. We further analyze the robustness-accuracy trade-off in \cref{sec:more}. Note that TGA-ZSR relies on patch-level attention maps in ViT models and is therefore not applicable to ResNet-based architectures.

\noindent\textbf{Out-of-Distribution Performance.}  
We further evaluate out-of-distribution generalization on two ImageNet variants: ImageNet-R \cite{hendrycks2021many} and ImageNet-S \cite{wang2019learning}. The results are summarized in \cref{tab:7}. Since ImageNet is used for adversarial fine-tuning, we treat it as an in-domain reference and focus primarily on the two shifted variants. AGFT achieves the best clean and robust accuracy on both ImageNet-R and ImageNet-S. While TGA-ZSR performs better on ImageNet itself, AGFT generalizes more effectively to the distribution-shifted variants, demonstrating stronger out-of-distribution robustness and generalization.

\begin{table}[t]
\caption{\textbf{Out-of-distribution} performance on ImageNet variants under PGD-20 attack
with perturbation budget $\epsilon = 1/255$.}
\vspace{-9pt} 
\centering
\footnotesize 
\setlength{\tabcolsep}{4pt} 
\renewcommand{\arraystretch}{0.9}
\begin{tabular}{c c c c c c c}
\toprule
\multirow{2}{*}[-0.6ex]{Method} & \multicolumn{2}{c}{ImageNet} & \multicolumn{2}{c}{ImageNet-R} & \multicolumn{2}{c}{ImageNet-S} \\
\cmidrule(lr){2-3} \cmidrule(lr){4-5} \cmidrule(lr){6-7}
& Clean & Robust & Clean & Robust & Clean & Robust\\
\midrule
TeCoA       & 63.50 & 41.29 & 55.62 & 37.25 & 34.41  & 22.92 \\
PMG-AFT      & 53.57 & 35.05 & 59.33 & 41.08 & 35.38  & 24.07 \\
TGA-ZSR       & 70.96  & 52.80 & 57.61 & 40.74 & 38.50  & 27.49 \\
\rowcolor{tableblue}
\textbf{AGFT (Ours)} & {60.11} & {44.95} & \textbf{61.70}  & \textbf{45.21} & \textbf{39.58} & \textbf{29.63} \\
\bottomrule
\end{tabular}
\label{tab:7}
\end{table}

\begin{table}[t]
\caption{\textbf{Ablation study} of hyperparameters.
The original pre-trained CLIP setting corresponds to $(\gamma = 1.0, \tau = 1/100)$. Entries are reported as clean accuracy / robust accuracy.}
\vspace{-9pt} 
\renewcommand{\arraystretch}{0.9}
\centering
\footnotesize 
\setlength{\tabcolsep}{4pt} 
\begin{tabular}{ccccc}
\toprule
$1/\tau$ & {100} & {140} & {180} & {220} \\
\midrule
{$\gamma=1.0$}   & \textbf{60.97 / 45.07} & 59.78 / 45.60 & 58.96 / 45.81 & 58.36 / 45.97 \\
{$\gamma=0.8$}     & 61.65 / 44.94 & 60.49 / 45.87 & 59.53 / 46.19 & 58.79 / 46.25 \\
{$\gamma=0.6$}     & 62.05 / 44.72 & 61.31 / 46.11 & 60.39 / 46.57 & 59.55 / 46.79 \\
{$\gamma=0.4$}     & 62.20 / 43.17 & 61.90 / 45.41 & \textbf{61.35 / 46.57} & 60.68 / 47.08 \\
{$\gamma=0.2$}     & 63.23 / 33.29 & 62.19 / 40.41 & 62.00 / 43.00 & 61.77 / 44.38 \\
\bottomrule
\end{tabular}
\vspace{-6pt} 
\label{tab:8}
\end{table}

\subsection{Ablation Study}
\label{sec:ablation}

We conduct an ablation study on the two key hyperparameters in AGFT, $\gamma$ and $1/\tau$, and report the clean and robust accuracy in \cref{tab:8}. As $\gamma$ decreases, clean accuracy improves, indicating better alignment with the robust feature space. However, robust accuracy peaks at an intermediate $\gamma$: moderate calibration enhances robustness, while an overly small $\gamma$ leads to over-smoothing of the target distribution, making the model more sensitive to adversarial perturbations. 

Increasing $1/\tau$ improves robust accuracy but gradually reduces clean accuracy, as the resulting sharper targets closely resemble hard labels, which strengthen adversarial supervision at the cost of cross-modal semantic preservation. Overall, AGFT strikes a favorable balance between robust feature adaptation and semantic structure preservation, improving both clean and robust accuracy.

Although \cref{tab:8} presents a grid search, exhaustive tuning is unnecessary. In practice, we first tune $\gamma$ to set the alignment scale, then adjust $1/\tau$ to refine the robustness-accuracy trade-off. For example, the path $(1.0,100) \to (0.6,100) \to (0.6,140) \to (0.4,140) \to (0.4,180)$ provides an efficient linear search procedure.

\begin{figure}[t]
\vspace{-12pt}
  \centering
   \includegraphics[width=0.9\linewidth]{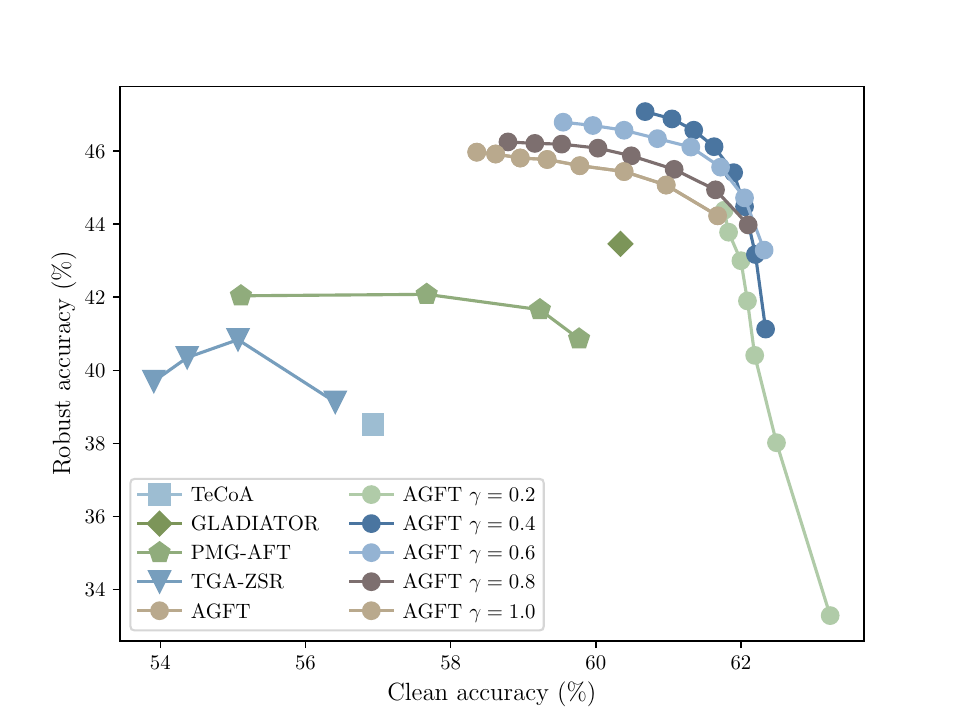}
   \vspace{-7pt}
   \caption{\textbf{Trade-off} between robust and clean accuracy across different methods. Each marker type denotes one method, and each point corresponds to a different trade-off configuration.}
   \label{fig:3}
\vspace{-9pt}
\end{figure}

\subsection{More Analysis}
\label{sec:more}
\noindent\textbf{Trade-off between Robustness and Accuracy.}  
Based on the results in \cref{tab:8}, we further analyze how different adversarial fine-tuning methods influence the trade-off between clean accuracy and adversarial robustness. As shown in \cref{fig:3}, we plot the average clean and robust accuracy across the 15 datasets under different trade-off configurations. By adjusting its hyperparameters, AGFT offers flexible control over the trade-off. Compared to the baselines, AGFT achieves a superior Pareto frontier, consistently delivering a better robustness-accuracy trade-off across a broad range of operating points. Additional baseline configurations can be found in the Appendix (\cref{sec:baselines}).

\noindent\textbf{Empirical Study of Distribution Consistency Calibration.}  
We calibrate the original output distribution by attenuating the confidence scale while preserving the intrinsic cross-modal semantic structure. To assess its impact, we report the output confidence along with the maximum and mean cross-modal similarity scores between each image and the candidate texts in \cref{tab:9}. Compared to the original CLIP model, all adversarially fine-tuned models show reduced confidence and lower similarity scores, indicating that adversarial fine-tuning naturally alters the confidence scale and similarity structure of the pre-trained model. Such empirical evidence reinforces the design of AGFT, which explicitly calibrates the target distribution rather than forcing the robust model to match the original confidence scale.

\begin{table}[t]
\caption{\textbf{Empirical study} of distribution consistency calibration. $conf$, $s_{max}$, and $s_{mean}$ represent the prediction confidence, and the maximum and mean cosine similarity, respectively ($\times 100$).}
\vspace{-9pt} 
\centering
\footnotesize 
\setlength{\tabcolsep}{4pt} 
\renewcommand{\arraystretch}{0.9}
\begin{tabular}{c c c c c c c}
\toprule
\multirow{2}{*}[-0.6ex]{Method} & \multicolumn{3}{c}{clean images} & \multicolumn{3}{c}{adversarial examples} \\
\cmidrule(lr){2-4} \cmidrule(lr){5-7}
& $conf$ & $s_{max}$ & $s_{mean}$ & $conf$ & $s_{max}$ & $s_{mean}$\\
\midrule
CLIP       & 63.51 & 29.19 & 21.66 & - & -  & - \\
TeCoA       & 54.15 & 28.11 & 20.82 & 50.68 & 27.85  & 20.85 \\
PMG-AFT      & 53.81 & 28.77 & 22.02 & 49.85 & 28.53  & 22.04 \\
TGA-ZSR       & 59.98  & 26.64 & 18.13 & 58.15 & 26.48  & 18.16 \\
\rowcolor{tableblue}
\textbf{AGFT (Ours)} & {23.72} & {17.96} & {15.02}  & {20.75} & {17.75} & {15.04} \\
\bottomrule
\end{tabular}
\label{tab:9}
\end{table}

\begin{table}[t]
\caption{\textbf{Computational cost} comparison: numbers of forward and backward propagation, and training time per epoch (min).}
\vspace{-9pt} 
\centering
\footnotesize 
\setlength{\tabcolsep}{5pt} 
\renewcommand{\arraystretch}{0.9} 
\begin{tabular}{c c c c c}
\toprule
{} & TeCoA & PMG-AFT & TGA-ZSR & {\cellcolor{tableblue}\textbf{AGFT (Ours)}} \\
\midrule
Forward & 1 & 3 & 3 & \cellcolor{tableblue}2 \\
Backward & 1 & 2 & 2 & \cellcolor{tableblue}1 \\
Time & 23.45 & 34.36 & 39.78 & \cellcolor{tableblue}25.96 \\
\bottomrule
\end{tabular}%
\vspace{-6pt} 
\label{tab:10}
\end{table}

\noindent\textbf{Preservation of Semantic Structure.}
Zero-shot generalization in VLMs relies on a continuous semantic space where relative similarities to non-target concepts encode fine-grained semantics. Conventional classification-guided adversarial training uses hard labels, which severely penalize non-target similarities and distort the original relational topology. By contrast, AGFT leverages pre-trained soft distributions as anchors, maintaining the representation integrity essential for zero-shot transfer.

\begin{figure}[t]
  \vspace{-6pt}
  \centering
  \begin{minipage}{0.32\linewidth} 
    \begin{subfigure}{\linewidth}
    \centering 
    \includegraphics[width=0.9\linewidth]{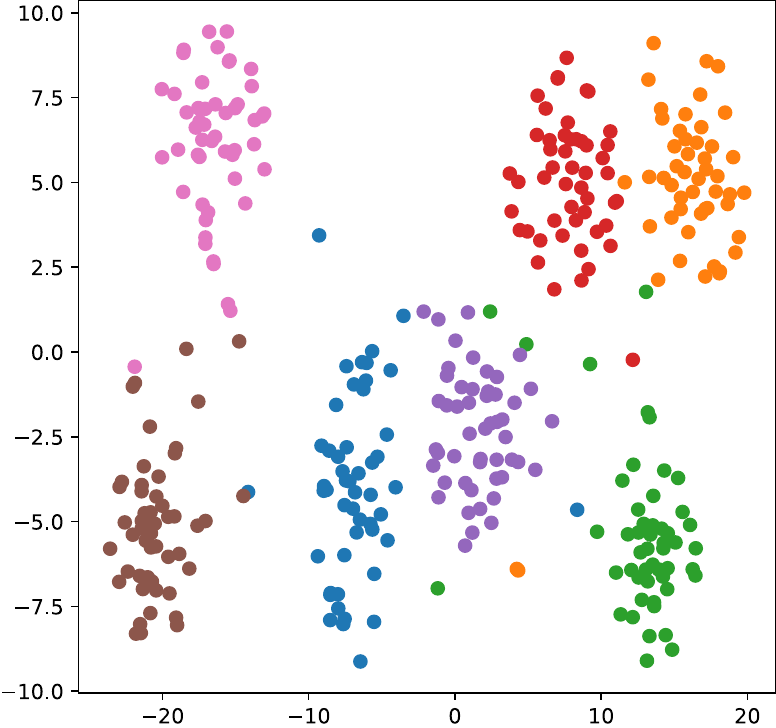}
    \vspace{-2.5pt}
    \caption{Pre-trained}
    \end{subfigure}
  \end{minipage}
  \begin{minipage}{0.32\linewidth} 
    \begin{subfigure}{\linewidth}
    \centering 
    \includegraphics[width=0.9\linewidth]{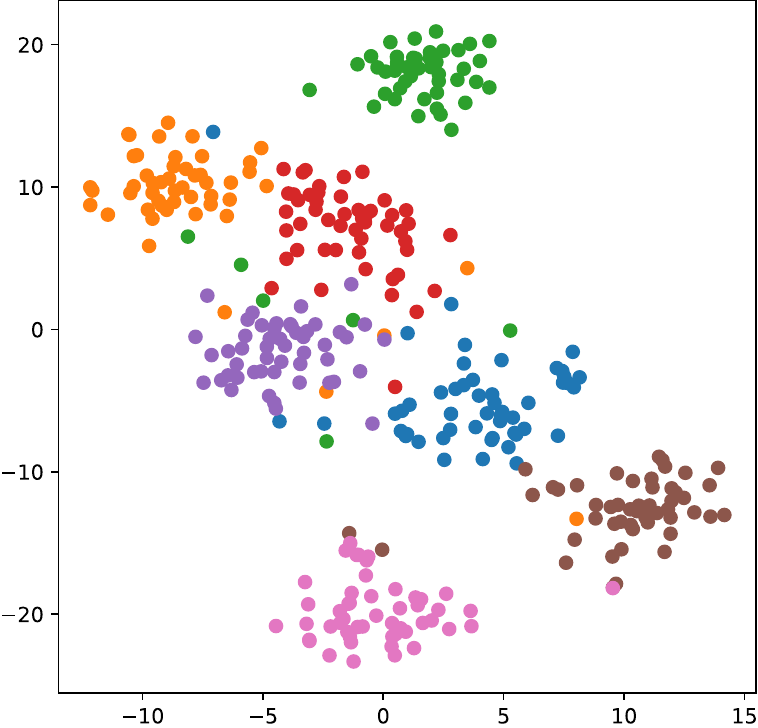}
    \vspace{-2.5pt}
    \caption{AGFT}
    \end{subfigure}
  \end{minipage}
  \begin{minipage}{0.32\linewidth} 
    \begin{subfigure}{\linewidth}
    \centering 
    \includegraphics[width=0.9\linewidth]{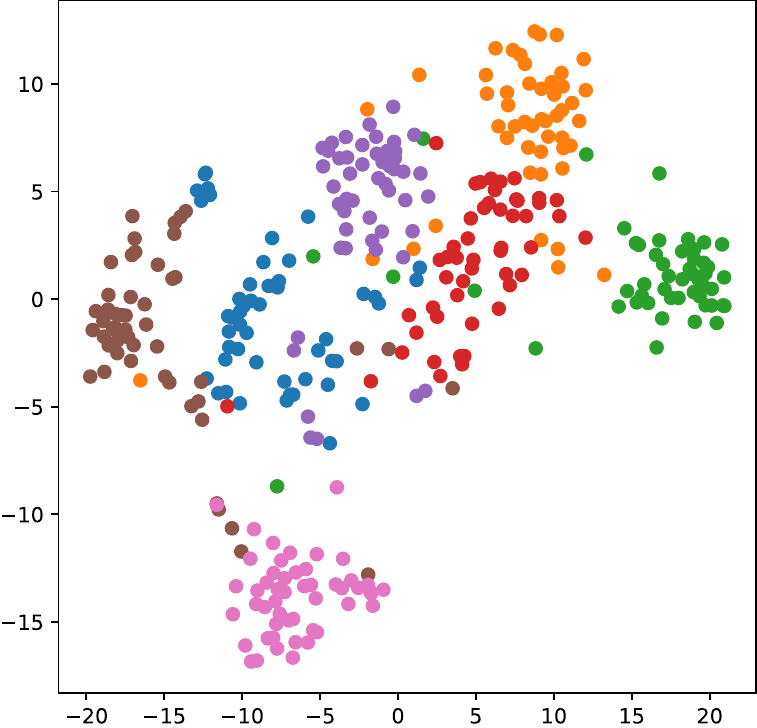}
    \vspace{-2.5pt}
    \caption{TeCoA}
    \end{subfigure}
  \end{minipage}
  \vspace{-7pt}
  \caption{
    T-SNE visualization of 7 image categories.
    (a) The original pre-trained CLIP evaluated on clean images.
    Both (b) AGFT and (c) TeCoA are evaluated on adversarial examples.
    }
  \label{fig:4}
  \vspace{-10pt}
\end{figure}

We validate this structural preservation both quantitatively and qualitatively. First, we assess the Top-5 prediction overlap (IoU) between the fine-tuned and original pre-trained models.
This metric calculates the Intersection-over-Union of the Top-5 predicted classes per image, which is then averaged over the test set. Notably, the original model is always evaluated on clean images to serve as a fixed baseline, while the fine-tuned model is evaluated on clean and adversarial images separately. AGFT effectively prevents the logit collapse typical of hard-label methods, retaining a high distribution overlap of $62.48\%$ on clean images and $56.12\%$ on adversarial examples, markedly outperforming the baseline TeCoA ($48.01\%$ and $42.98\%$, respectively). Second, the t-SNE visualization presented in \cref{fig:4} provides qualitative evidence of feature space preservation. While adversarial perturbations induce pronounced structural distortion and cluster overlapping in TeCoA, AGFT maintains a coherent feature organization that closely mirrors the original pre-trained semantic neighborhoods.

\noindent\textbf{Computational Efficiency.}  
In addition to robustness and generalization, computational efficiency is an important consideration. We compare AGFT with prior methods in \cref{tab:10}. Excluding the uniform PGD-2 generation overhead shared across all baselines, we report solely the forward and backward propagation steps required for the subsequent model updates. TeCoA directly optimizes the model on adversarial examples, resulting in the lowest training cost. PMG-AFT and TGA-ZSR introduce additional supervision from clean images and pre-trained models, requiring extra forward and backward passes. TGA-ZSR is particularly costly due to the computation of attention maps. In contrast, AGFT only requires one additional forward pass to obtain the pre-trained output distribution. As a result, AGFT achieves strong performance gains while maintaining favorable computational efficiency.

\section{Conclusion}
\label{sec:conclusion}
In this paper, we propose AGFT, a novel alignment-guided adversarial fine-tuning framework designed to enhance the zero-shot adversarial robustness of vision-language models. By replacing label-based supervision with the probabilistic predictions of the pre-trained model, AGFT effectively preserves the pre-trained cross-modal alignment during adversarial fine-tuning. Additionally, the proposed distribution consistency calibration ensures the integrity of the visual-textual correspondence structure. Extensive experiments across diverse zero-shot benchmarks, attack settings, perturbation budgets, architectures, and out-of-distribution scenarios demonstrate that AGFT consistently outperforms existing methods in both robustness and clean accuracy, all while maintaining computational efficiency.

{
    \small
    \bibliographystyle{ieeenat_fullname}
    \bibliography{main}

@String(CVPR= {IEEE Conf. Comput. Vis. Pattern Recog.})

@String(ICCV= {Int. Conf. Comput. Vis.})

@String(ECCV= {Eur. Conf. Comput. Vis.})

@String(NIPS= {Adv. Neural Inform. Process. Syst.})

@String(ACMMM= {ACM Int. Conf. Multimedia})

@String(ICLR = {Int. Conf. Learn. Represent.})

@String(AAAI = {AAAI})

@String(CVPRW= {IEEE Conf. Comput. Vis. Pattern Recog. Worksh.})

@String(CVPR  = {CVPR})

@String(ICCV  = {ICCV})

@String(ECCV  = {ECCV})

@String(NIPS  = {NeurIPS})

@String(ACMMM = {ACM MM})

@String(ICLR  = {ICLR})

@String(CVPRW= {CVPRW})

@inproceedings{jia2021scaling,
  title={Scaling up visual and vision-language representation learning with noisy text supervision},
  author={Jia, Chao and Yang, Yinfei and Xia, Ye and Chen, Yi-Ting and Parekh, Zarana and Pham, Hieu and Le, Quoc and Sung, Yun-Hsuan and Li, Zhen and Duerig, Tom},
  booktitle={ICML},
  pages={4904--4916},
  year={2021},
}

@inproceedings{li2022blip,
  title={Blip: Bootstrapping language-image pre-training for unified vision-language understanding and generation},
  author={Li, Junnan and Li, Dongxu and Xiong, Caiming and Hoi, Steven},
  booktitle={ICML},
  pages={12888--12900},
  year={2022},
}

@article{li2025perception,
  title={Perception, reason, think, and plan: A survey on large multimodal reasoning models},
  author={Li, Yunxin and Liu, Zhenyu and Li, Zitao and Zhang, Xuanyu and Xu, Zhenran and Chen, Xinyu and Shi, Haoyuan and Jiang, Shenyuan and Wang, Xintong and Wang, Jifang and others},
  journal={arXiv preprint arXiv:2505.04921},
  year={2025}
}

@inproceedings{radford2021learning,
  title={Learning transferable visual models from natural language supervision},
  author={Radford, Alec and Kim, Jong Wook and Hallacy, Chris and Ramesh, Aditya and Goh, Gabriel and Agarwal, Sandhini and Sastry, Girish and Askell, Amanda and Mishkin, Pamela and Clark, Jack and others},
  booktitle={ICML},
  pages={8748--8763},
  year={2021},
}

@inproceedings{szegedy2013intriguing,
  title={Intriguing properties of neural networks},
  author={Szegedy, Christian and Zaremba, Wojciech and Sutskever, Ilya and Bruna, Joan and Erhan, Dumitru and Goodfellow, Ian and Fergus, Rob},
  booktitle=ICLR,
  year={2014}
}

@inproceedings{goodfellow2014explaining,
  title={Explaining and harnessing adversarial examples},
  author={Goodfellow, Ian J and Shlens, Jonathon and Szegedy, Christian},
  booktitle=ICLR,
  year={2015}
}

@inproceedings{li2025transferable,
  title={Transferable adversarial face attack with text controlled attribute},
  author={Li, Wenyun and Zhang, Zheng and Lan, Xiangyuan and Jiang, Dongmei},
  booktitle=AAAI,
  volume={39},
  number={5},
  pages={4977--4985},
  year={2025}
}

@inproceedings{zhao2023evaluating,
  title={On evaluating adversarial robustness of large vision-language models},
  author={Zhao, Yunqing and Pang, Tianyu and Du, Chao and Yang, Xiao and Li, Chongxuan and Cheung, Ngai-Man Man and Lin, Min},
  booktitle=NIPS,
  volume={36},
  year={2023}
}

@inproceedings{zhou2022extract,
  title={Extract free dense labels from clip},
  author={Zhou, Chong and Loy, Chen Change and Dai, Bo},
  booktitle=ECCV,
  pages={696--712},
  year={2022},
}

@inproceedings{madry2017towards,
  title={Towards deep learning models resistant to adversarial attacks},
  author={Madry, Aleksander and Makelov, Aleksandar and Schmidt, Ludwig and Tsipras, Dimitris and Vladu, Adrian},
  booktitle=ICLR,
  year={2018}
}

@inproceedings{chen2022visualgpt,
  title={VisualGPT: Data-efficient adaptation of pretrained language models for image captioning},
  author={Chen, Jun and Guo, Han and Yi, Kai and Li, Boyang and Elhoseiny, Mohamed},
  booktitle=CVPR,
  pages={18030--18040},
  year={2022}
}

@inproceedings{mao2022understanding,
  title={Understanding zero-shot adversarial robustness for large-scale models},
  author={Mao, Chengzhi and Geng, Scott and Yang, Junfeng and Wang, Xin and Vondrick, Carl},
  booktitle=ICLR,
  year={2023}
}

@inproceedings{wang2024pre,
  title={Pre-trained model guided fine-tuning for zero-shot adversarial robustness},
  author={Wang, Sibo and Zhang, Jie and Yuan, Zheng and Shan, Shiguang},
  booktitle=CVPR,
  pages={24502--24511},
  year={2024}
}

@inproceedings{yu2024text,
  title={Text-guided attention is all you need for zero-shot robustness in vision-language models},
  author={Yu, Lu and Zhang, Haiyang and Xu, Changsheng},
  booktitle=NIPS,
  volume={37},
  year={2024}
}

@inproceedings{dong2025stabilizing,
  title={Stabilizing modality gap \& lowering gradient norms improve zero-shot adversarial robustness of vlms},
  author={Dong, Junhao and Koniusz, Piotr and Qu, Xinghua and Ong, Yew-Soon},
  booktitle={SIGKDD},
  pages={236--247},
  year={2025}
}

@inproceedings{deng2009imagenet,
  title={ImageNet: A large-scale hierarchical image database},
  author={Deng, Jia and Dong, Wei and Socher, Richard and Li, Li-Jia and Li, Kai and Fei-Fei, Li},
  booktitle=CVPR,
  pages={248--255},
  year={2009},
}

@inproceedings{carlini2017towards,
  title={Towards evaluating the robustness of neural networks},
  author={Carlini, Nicholas and Wagner, David},
  booktitle={2017 IEEE symposium on security and privacy (SP)},
  pages={39--57},
  year={2017},
}

@inproceedings{moosavi2016deepfool,
  title={DeepFool: a simple and accurate method to fool deep neural networks},
  author={Moosavi-Dezfooli, Seyed-Mohsen and Fawzi, Alhussein and Frossard, Pascal},
  booktitle=CVPR,
  pages={2574--2582},
  year={2016}
}

@inproceedings{croce2020reliable,
  title={Reliable evaluation of adversarial robustness with an ensemble of diverse parameter-free attacks},
  author={Croce, Francesco and Hein, Matthias},
  booktitle={ICML},
  pages={2206--2216},
  year={2020},
}

@inproceedings{metzen2017detecting,
  title={On detecting adversarial perturbations},
  author={Metzen, Jan Hendrik and Genewein, Tim and Fischer, Volker and Bischoff, Bastian},
  booktitle=ICLR,
  year={2017}
}

@inproceedings{nie2022diffusion,
  title={Diffusion Models for Adversarial Purification},
  author={Nie, Weili and Guo, Brandon and Huang, Yujia and Xiao, Chaowei and Vahdat, Arash and Anandkumar, Animashree},
  booktitle={ICML},
  pages={16805--16827},
  year={2022},
}

@inproceedings{papernot2016distillation,
  title={Distillation as a defense to adversarial perturbations against deep neural networks},
  author={Papernot, Nicolas and McDaniel, Patrick and Wu, Xi and Jha, Somesh and Swami, Ananthram},
  booktitle={2016 IEEE symposium on security and privacy (SP)},
  pages={582--597},
  year={2016},
}

@inproceedings{athalye2018obfuscated,
  title={Obfuscated gradients give a false sense of security: Circumventing defenses to adversarial examples},
  author={Athalye, Anish and Carlini, Nicholas and Wagner, David},
  booktitle={ICML},
  pages={274--283},
  year={2018},
}

@inproceedings{shafahi2019adversarial,
  title={Adversarial training for free!},
  author={Shafahi, Ali and Najibi, Mahyar and Ghiasi, Mohammad Amin and Xu, Zheng and Dickerson, John and Studer, Christoph and Davis, Larry S and Taylor, Gavin and Goldstein, Tom},
  booktitle=NIPS,
  volume={32},
  year={2019}
}

@inproceedings{wong2020fast,
  title={Fast is better than free: Revisiting adversarial training},
  author={Wong, Eric and Rice, Leslie and Kolter, J Zico},
  booktitle=ICLR,
  year={2020}
}

@inproceedings{zhang2019theoretically,
  title={Theoretically principled trade-off between robustness and accuracy},
  author={Zhang, Hongyang and Yu, Yaodong and Jiao, Jiantao and Xing, Eric and El Ghaoui, Laurent and Jordan, Michael},
  booktitle={ICML},
  pages={7472--7482},
  year={2019},
}

@inproceedings{wang2019improving,
  title={Improving adversarial robustness requires revisiting misclassified examples},
  author={Wang, Yisen and Zou, Difan and Yi, Jinfeng and Bailey, James and Ma, Xingjun and Gu, Quanquan},
  booktitle=ICLR,
  year={2019}
}

@inproceedings{li2024one,
  title={One prompt word is enough to boost adversarial robustness for pre-trained vision-language models},
  author={Li, Lin and Guan, Haoyan and Qiu, Jianing and Spratling, Michael},
  booktitle=CVPR,
  pages={24408--24419},
  year={2024}
}

@inproceedings{zhou2025parameter,
  title={Parameter-free and Accessible Prompt Learning to Enhance Adversarial Robustness for Pre-trained Vision-Language Models},
  author={Zhou, Xingran and Yang, Kun and Miao, Changtao and Hu, Bingyu and Xu, Zhuoer and Cui, Shiwen and Meng, Changhua and Hong, Dan},
  booktitle={NAACL},
  pages={751--761},
  year={2025}
}

@inproceedings{zhang2024adversarial,
  title={Adversarial prompt tuning for vision-language models},
  author={Zhang, Jiaming and Ma, Xingjun and Wang, Xin and Qiu, Lingyu and Wang, Jiaqi and Jiang, Yu-Gang and Sang, Jitao},
  booktitle=ECCV,
  pages={56--72},
  year={2024},
}

@inproceedings{sheng2025r,
  title={R-TPT: Improving Adversarial Robustness of Vision-Language Models through Test-Time Prompt Tuning},
  author={Sheng, Lijun and Liang, Jian and Wang, Zilei and He, Ran},
  booktitle=CVPR,
  pages={29958--29967},
  year={2025}
}

@inproceedings{tong2025zero,
  title={On the Zero-shot Adversarial Robustness of Vision-Language Models: A Truly Zero-shot and Training-free Approach},
  author={Tong, Baoshun and Lai, Hanjiang and Pan, Yan and Yin, Jian},
  booktitle=CVPR,
  pages={19921--19930},
  year={2025}
}

@inproceedings{wang2025tapt,
  title={Tapt: Test-time adversarial prompt tuning for robust inference in vision-language models},
  author={Wang, Xin and Chen, Kai and Zhang, Jiaming and Chen, Jingjing and Ma, Xingjun},
  booktitle=CVPR,
  pages={19910--19920},
  year={2025}
}

@inproceedings{xing2025clip,
  title={CLIP is strong enough to fight back: Test-time counterattacks towards zero-shot adversarial robustness of clip},
  author={Xing, Songlong and Zhao, Zhengyu and Sebe, Nicu},
  booktitle=CVPR,
  pages={15172--15182},
  year={2025}
}

@inproceedings{fei2004learning,
  title={Learning generative visual models from few training examples: An incremental bayesian approach tested on 101 object categories},
  author={Fei-Fei, Li and Fergus, Rob and Perona, Pietro},
  booktitle=CVPRW,
  pages={178--178},
  year={2004},
}

@techreport{griffin2007caltech,
  title={Caltech-256 object category dataset},
  author={Griffin, Gregory and Holub, Alex and Perona, Pietro and others},
  year={2007},
  institution={Technical Report 7694, California Institute of Technology Pasadena}
}

@article{krizhevsky2009learning,
  title={Learning multiple layers of features from tiny images},
  author={Krizhevsky, Alex and others},
  year={2009}
}

@inproceedings{coates2011analysis,
  title={An analysis of single-layer networks in unsupervised feature learning},
  author={Coates, Adam and Ng, Andrew and Lee, Honglak},
  booktitle={Proceedings of the fourteenth international conference on artificial intelligence and statistics},
  pages={215--223},
  year={2011},
}

@techreport{maji2013fine,
   title         = {Fine-Grained Visual Classification of Aircraft},
   author        = {Maji, Subhransu and Rahtu, Esa and Kannala, Juho and Blaschko, Matthew and Vedaldi, Andrea},
   year          = {2013},
   archivePrefix = {arXiv},
   eprint        = {1306.5151},
   primaryClass  = "cs-cv",
}

@inproceedings{nilsback2008automated,
  title={Automated flower classification over a large number of classes},
  author={Nilsback, Maria-Elena and Zisserman, Andrew},
  booktitle={2008 Sixth Indian conference on computer vision, graphics \& image processing},
  pages={722--729},
  year={2008},
}

@inproceedings{bossard2014food,
  title={Food-101--Mining discriminative components with random forests},
  author={Bossard, Lukas and Guillaumin, Matthieu and Van Gool, Luc},
  booktitle=ECCV,
  pages={446--461},
  year={2014},
}

@inproceedings{parkhi2012cats,
  title={Cats and dogs},
  author={Parkhi, Omkar M and Vedaldi, Andrea and Zisserman, Andrew and Jawahar, CV},
  booktitle=CVPR,
  pages={3498--3505},
  year={2012},
}

@inproceedings{krause20133d,
  title={3d object representations for fine-grained categorization},
  author={Krause, Jonathan and Stark, Michael and Deng, Jia and Fei-Fei, Li},
  booktitle={Proceedings of the IEEE international conference on computer vision workshops},
  pages={554--561},
  year={2013}
}

@inproceedings{xiao2010sun,
  title={SUN database: Large-scale scene recognition from abbey to zoo},
  author={Xiao, Jianxiong and Hays, James and Ehinger, Krista A and Oliva, Aude and Torralba, Antonio},
  booktitle={2010 IEEE computer society conference on computer vision and pattern recognition},
  pages={3485--3492},
  year={2010},
}

@inproceedings{cimpoi2014describing,
  title={Describing textures in the wild},
  author={Cimpoi, Mircea and Maji, Subhransu and Kokkinos, Iasonas and Mohamed, Sammy and Vedaldi, Andrea},
  booktitle=CVPR,
  pages={3606--3613},
  year={2014}
}

@article{helber2019eurosat,
  title={EuroSAT: A novel dataset and deep learning benchmark for land use and land cover classification},
  author={Helber, Patrick and Bischke, Benjamin and Dengel, Andreas and Borth, Damian},
  journal={IEEE Journal of Selected Topics in Applied Earth Observations and Remote Sensing},
  volume={12},
  number={7},
  pages={2217--2226},
  year={2019},
  publisher={IEEE}
}

@inproceedings{veeling2018rotation,
  title={Rotation equivariant CNNs for digital pathology},
  author={Veeling, Bastiaan S and Linmans, Jasper and Winkens, Jim and Cohen, Taco and Welling, Max},
  booktitle={International Conference on Medical image computing and computer-assisted intervention},
  pages={210--218},
  year={2018},
}

@inproceedings{hendrycks2021many,
  title={The many faces of robustness: A critical analysis of out-of-distribution generalization},
  author={Hendrycks, Dan and Basart, Steven and Mu, Norman and Kadavath, Saurav and Wang, Frank and Dorundo, Evan and Desai, Rahul and Zhu, Tyler and Parajuli, Samyak and Guo, Mike and others},
  booktitle=ICCV,
  pages={8340--8349},
  year={2021}
}

@inproceedings{wang2019learning,
  title={Learning robust global representations by penalizing local predictive power},
  author={Wang, Haohan and Ge, Songwei and Lipton, Zachary and Xing, Eric P},
  booktitle=NIPS,
  volume={32},
  year={2019}
}

@inproceedings{dong2018boosting,
  title={Boosting adversarial attacks with momentum},
  author={Dong, Yinpeng and Liao, Fangzhou and Pang, Tianyu and Su, Hang and Zhu, Jun and Hu, Xiaolin and Li, Jianguo},
  booktitle=CVPR,
  pages={9185--9193},
  year={2018}
}

@inproceedings{xie2019improving,
  title={Improving transferability of adversarial examples with input diversity},
  author={Xie, Cihang and Zhang, Zhishuai and Zhou, Yuyin and Bai, Song and Wang, Jianyu and Ren, Zhou and Yuille, Alan L},
  booktitle=CVPR,
  pages={2730--2739},
  year={2019}
}

@inproceedings{dong2019evading,
  title={Evading defenses to transferable adversarial examples by translation-invariant attacks},
  author={Dong, Yinpeng and Pang, Tianyu and Su, Hang and Zhu, Jun},
  booktitle=CVPR,
  pages={4312--4321},
  year={2019}
}

@inproceedings{lin2019nesterov,
  title={Nesterov accelerated gradient and scale invariance for adversarial attacks},
  author={Lin, Jiadong and Song, Chuanbiao and He, Kun and Wang, Liwei and Hopcroft, John E},
  booktitle=ICLR,
  year={2020}
}

@inproceedings{gao2020patch,
  title={Patch-wise attack for fooling deep neural network},
  author={Gao, Lianli and Zhang, Qilong and Song, Jingkuan and Liu, Xianglong and Shen, Heng Tao},
  booktitle=ECCV,
  pages={307--322},
  year={2020},
}

@article{gao2020patch2,
  title={Patch-wise++ perturbation for adversarial targeted attacks},
  author={Gao, Lianli and Zhang, Qilong and Song, Jingkuan and Shen, Heng Tao},
  journal={arXiv preprint arXiv:2012.15503},
  year={2020}
}

@inproceedings{dosovitskiy2020image,
  title={An image is worth 16x16 Words: Transformers for image recognition at scale},
  author={Dosovitskiy, Alexey and Beyer, Lucas and Kolesnikov, Alexander and Weissenborn, Dirk and Zhai, Xiaohua and Unterthiner, Thomas and Dehghani, Mostafa and Minderer, Matthias and Heigold, Georg and Gelly, Sylvain and others},
  booktitle=ICLR,
  year={2021}
}

@inproceedings{he2016deep,
  title={Deep residual learning for image recognition},
  author={He, Kaiming and Zhang, Xiangyu and Ren, Shaoqing and Sun, Jian},
  booktitle=CVPR,
  pages={770--778},
  year={2016}
}

@inproceedings{bhagwatkar2024improving,
  title={Improving adversarial robustness in vision-language models with architecture and prompt design},
  author={Bhagwatkar, Rishika and Nayak, Shravan and Bashivan, Pouya and Rish, Irina},
  booktitle={Findings of EMNLP},
  pages={17003--17020},
  year={2024}
}

@inproceedings{li2020bert,
  title={BERT-ATTACK: Adversarial attack against BERT using BERT},
  author={Li, Linyang and Ma, Ruotian and Guo, Qipeng and Xue, Xiangyang and Qiu, Xipeng},
  booktitle={EMNLP},
  pages={6193--6202},
  year={2020}
}

@inproceedings{zhang2022towards,
  title={Towards adversarial attack on vision-language pre-training models},
  author={Zhang, Jiaming and Yi, Qi and Sang, Jitao},
  booktitle=ACMMM,
  pages={5005--5013},
  year={2022}
}

@inproceedings{mengpolaformer,
  title={PolaFormer: Polarity-aware Linear Attention for Vision Transformers},
  author={Meng, Weikang and Luo, Yadan and Li, Xin and Jiang, Dongmei and Zhang, Zheng},
  booktitle=ICLR,
  year={2025}
}

@inproceedings{liu2023visual,
  title={Visual instruction tuning},
  author={Liu, Haotian and Li, Chunyuan and Wu, Qingyang and Lee, Yong Jae},
  booktitle=NIPS,
  volume={36},
  pages={34892--34916},
  year={2023}
}
}

\clearpage
\setcounter{page}{1}
\maketitlesupplementary

\section{Implementation Details of Baselines}
\label{sec:baselines}
In this section, we briefly describe the baseline methods and summarize their implementation details.
\begin{itemize}
\item TeCoA \cite{mao2022understanding} introduces a text-guided contrastive adversarial training objective and improves adversarial robustness by fine-tuning the CLIP image encoder. We follow the official implementation and hyperparameter settings.

\item PMG-AFT \cite{wang2024pre} extends TeCoA by leveraging supervision from the pre-trained model to alleviate overfitting. Because the original method was developed on a smaller-scale dataset rather than ImageNet, its default hyperparameters may not be optimal in our setting. We therefore tune only the learning rate and report the best-performing configuration in \cref{tab:11}.

\item TGA-ZSR \cite{yu2024text} enhances adversarial robustness through an attention refinement module and an attention-based model constraint module. Similar to PMG-AFT, we tune the learning rate to obtain a stronger configuration for our setting, as reported in \cref{tab:11}.

\item GLADIATOR \cite{dong2025stabilizing} improves robustness by maximizing the effective rank of features and applying feature noise modulation. As the official code is not currently accessible, we adopt the performance reported by the authors in their publication.
\end{itemize}

\section{Attack Configurations}
\label{sec:attack}
In addition to PGD \cite{madry2017towards}, we evaluate eight additional attacks, which fall into four categories:

\begin{itemize}
\item \textbf{Strong Attacks}:
Carlini \& Wagner (C\&W) attack \cite{carlini2017towards} and AutoAttack (AA) \cite{croce2020reliable}.

\item \textbf{Optimization-based Attacks}:
MI-FGSM \cite{dong2018boosting} and NI-FGSM \cite{lin2019nesterov}.

\item \textbf{Data-augmentation-based Attacks}:
DI\textsuperscript{2}-FGSM \cite{xie2019improving} and TI-FGSM \cite{dong2019evading}.

\item \textbf{Patch-wise Attacks}:
PI-FGSM \cite{gao2020patch} and PI-FGSM++ \cite{gao2020patch2}.
\end{itemize}
To adapt the C\&W attack to the $\ell_\infty$-norm threat model,
we optimize the objective within a PGD framework, using a fixed step size and projecting the perturbation onto the feasible $\ell_\infty$ ball at each iteration. For AutoAttack, we use its two APGD variants, namely APGD-CE and APGD-DLR.

In addition to standard untargeted attacks, we also consider targeted attacks, where adversarial examples are optimized to be classified into a specified target class.
The target class is either selected uniformly at random or chosen as the least-likely class under the clean-image prediction.
We evaluate targeted versions of PGD \cite{madry2017towards}, MI-FGSM \cite{dong2018boosting}, NI-FGSM \cite{lin2019nesterov}, DI\textsuperscript{2}-FGSM \cite{xie2019improving}, and TI-FGSM \cite{dong2019evading}.

\begin{table}[t]
\caption{Average accuracy (\%) under different hyperparameters for PMG-AFT and TGA-ZSR.}
\vspace{-9pt} 
\renewcommand{\arraystretch}{0.9}
\centering
\footnotesize 

\begin{tabular}{c c c c}
\toprule
{Method} & {Learning Rate} & {Clean Acc.} & {Robust Acc.}\\
\midrule

\multirow{5}{*}{{PMG-AFT}} 
& $lr=$ 5e-6     & 59.77 & 40.86\\
& {$lr=$ 1e-5 (ours)}    & \textbf{59.23} & \textbf{41.66}\\
& $lr=$ 2e-5      & 57.67 & 42.08\\
& $lr=$ 4e-5     & 55.11 & 42.04\\
& $lr=$ 5e-5 (original)     & 53.87 & 41.54\\
\midrule

\multirow{4}{*}{{TGA-ZSR}} 
& $lr=$ 1e-5    & 56.41 & 39.13\\
& {$lr=$ 4e-5 (ours)}     & \textbf{55.07} & \textbf{40.84}\\
& $lr=$ 7e-5     & 54.37 & 40.35\\
& $lr=$ 1e-4 (original)     & 53.91 & 39.70\\
\bottomrule
\end{tabular}
\vspace{-6.5pt} 
\label{tab:11}
\end{table}

\section{Interpretation of Hyperparameter {$\gamma$}}
\label{sec:any}

Although $\gamma$ is introduced in the main paper as a temperature-scaling ratio on the calibrated target distribution, it also admits an equivalent similarity/logit-scaling interpretation. Specifically, the calibrated target distribution can be written as
\begin{equation}
\begin{split}
  p_{{rob}}^{i,j} &= \frac{\exp\left(\cos\left(f_{\theta_{{orig}}}(x^i), f_\phi(t^j)\right)/(\tau/\gamma)\right)}{\sum_k \exp\left(\cos\left(f_{\theta_{{orig}}}(x^i), f_\phi(t^k)\right)/(\tau/\gamma)\right)}, \\
  &= \frac{\exp\left(\gamma \cdot \cos\left(f_{\theta_{{orig}}}(x^i), f_\phi(t^j)\right)/\tau\right)}{\sum_k \exp\left(\gamma \cdot \cos\left(f_{\theta_{{orig}}}(x^i), f_\phi(t^k)\right)/\tau\right)}.
  \label{eq:11}
\end{split}
\end{equation}
Therefore, temperature scaling in the distribution space is equivalent to uniformly scaling the original similarities/logits before the softmax. In this section, we use this equivalent similarity-scaling view to provide an intuitive interpretation of the role of $\gamma$.

Given an image $x \in \mathcal{X}$ and a text $t \in \mathcal{T}$, let
$f_{\theta_{{orig}}}(x)$ be the pre-trained visual feature,
$f_{\theta_{{rob}}}(x)$ be the robust visual feature,
and $f_{\phi}(t)$ be the text feature.
We $\ell_2$-normalize these features as
\begin{equation}
\begin{aligned}
\alpha_{{orig}} &= {f_{\theta_{{orig}}}(x)}/{\|f_{\theta_{{orig}}}(x)\|_2},\\
\alpha_{{rob}}  &= {f_{\theta_{{rob}}}(x)}/{\|f_{\theta_{{rob}}}(x)\|_2},\\
\beta                &= {f_{\phi}(t)}/{\|f_{\phi}(t)\|_2}.
\end{aligned}
\label{eq:12}
\end{equation}
The normalized features satisfy:
\begin{equation}
\begin{aligned}
    \cos(f_{\theta_{orig}}(x),f_{\phi}(t)) &= {\langle \alpha_{orig}, \beta \rangle}, \\
    \cos(f_{\theta_{rob}}(x),f_{\phi}(t)) &= {\langle \alpha_{rob}, \beta \rangle}.
  \label{eq:13}
\end{aligned}
\end{equation}
We define $\delta=\alpha_{orig}-\alpha_{rob}$, and then:
\begin{equation}
    {\langle \alpha_{orig}, \beta \rangle} - {\langle \alpha_{rob}, \beta \rangle} = {\langle \delta, \beta \rangle}.
\label{eq:14}
\end{equation}
According to the Cauchy-Schwarz Inequality, we have:
\begin{equation}
  \begin{split}
    |{\langle \alpha_{orig}, \beta \rangle} - {\langle \alpha_{rob}, \beta \rangle}| &= |{\langle \delta, \beta \rangle}| \leq {\| \delta \|}_{2} \cdot {\| \beta \|}_{2} =  {\| \delta \|}_{2} \\
    |{\langle \alpha_{orig}, \beta \rangle} - {\langle \alpha_{rob}, \beta \rangle}| &\leq {\| \delta \|}_{2}.
  \end{split}
\label{eq:15}
\end{equation}
Temperature scaling softens the pre-trained distribution by reducing the effective logit scale. For semantically aligned prompts, the original similarity is typically positive, and adversarial fine-tuning often reduces its magnitude. We therefore consider the common case ${\langle \alpha_{rob}, \beta \rangle} \le {\langle \alpha_{orig}, \beta \rangle}$, which gives
\begin{equation}
  \begin{split}
    {\langle \alpha_{orig}, \beta \rangle} - {\langle \alpha_{rob}, \beta \rangle} &\leq {\| \delta \|}_{2} \\
    1 - \frac{\langle \alpha_{rob}, \beta \rangle}{\langle \alpha_{orig}, \beta \rangle} &\leq \frac{{\| \delta \|}_{2}}{\langle \alpha_{orig}, \beta \rangle} \\
    \rho = \frac {\langle \alpha _{rob}, \beta \rangle }{\langle \alpha _{orig}, \beta \rangle } &\geq 1 - \frac{{\| \delta \|}_{2}}{\langle \alpha_{orig}, \beta \rangle}.
  \end{split}
\label{eq:16}
\end{equation}
\cref{eq:16} lower-bounds the local similarity contraction factor $\rho$ in terms of the representation shift $\|\delta\|_2$: a larger shift leads to a smaller lower bound on $\rho$.

Other sign configurations yield analogous upper or lower bounds for $\rho$. In zero-shot vision-language classification, however, prediction is mainly determined by semantically aligned prompts with high positive similarity ($\langle \alpha_{orig}, \beta \rangle > 0$). Moreover, as shown in \cref{tab:9}, adversarial fine-tuning typically reduces the absolute scale of these peak cross-modal similarities. Therefore, the positive-similarity degradation case in \cref{eq:16} captures the dominant effect of representation shift.

The local factor $\rho$ is distinct from the global hyperparameter $\gamma$ used in AGFT. Rather, $\gamma$ can be viewed as a global approximation to the dominant contraction behavior over semantically aligned prompts, yielding a softer target distribution that matches the robust model. This discussion provides an intuitive explanation for why temperature calibration helps, rather than formally deriving the optimal $\gamma$.

\section{Additional Results}
\label{sec:additional}

We provide additional experimental results that complement those reported in the main text, including detailed robust accuracy on all 15 datasets.

\noindent\textbf{Robustness against Stronger Attacks.}
We report robust accuracy under the C\&W attack and AutoAttack. The detailed results are presented in \cref{tab:12} and \cref{tab:13}, respectively.

\noindent\textbf{Fine-tuning in Diverse Architectures.}
We further report robust accuracy for two additional CLIP backbones, namely ViT-B/16 and RN50$\times$4. The corresponding results are shown in \cref{tab:14} and \cref{tab:15}, respectively.

\section{Future Work}
\label{sec:Future}
Although AGFT demonstrates strong performance in improving the zero-shot adversarial robustness of vision-language models, our current evaluation is mainly focused on zero-shot image classification under the $\ell_\infty$ threat model. In future work, we envision expanding the proposed alignment-guided paradigm along three directions.

First, we will evaluate AGFT under a broader range of threat models, including $\ell_2$-bounded perturbations \cite{carlini2017towards}, text-level attacks \cite{li2020bert}, and joint text-image attacks \cite{zhang2022towards}. Second, we aim to generalize AGFT to a wider variety of model architectures, extending beyond CLIP to other cross-modal models \cite{liu2023visual} and transformer-based architectures \cite{mengpolaformer}. Finally, we intend to apply AGFT to a broader set of downstream multimodal tasks, such as visual question answering and image captioning, thereby contributing to the development of more secure and trustworthy foundation models.

\begin{table*}[t]
\captionsetup{justification=justified, singlelinecheck=false}
\caption{Zero-shot robust accuracy (\%).
Adversarial examples are generated by \textbf{C\&W} attack with
the perturbation budget $\epsilon = 1/255$.}
\vspace{-9pt} 
\centering
\resizebox{\linewidth}{!}{%
\renewcommand{\arraystretch}{0.9} 
\begin{tabular}{c c c c c c c c c c c c c c c c c}
\toprule
\textbf{Method} & \rotatebox{90}{Caltech101} & \rotatebox{90}{Caltech256} & \rotatebox{90}{CIFAR10} & \rotatebox{90}{CIFAR100} & \rotatebox{90}{DTD} & \rotatebox{90}{EuroSAT} & \rotatebox{90}{FGVC} & \rotatebox{90}{Flower102} & \rotatebox{90}{Food101} & \rotatebox{90}{ImageNet} & \rotatebox{90}{OxfordPet} & \rotatebox{90}{PCAM} & \rotatebox{90}{Stanf.Cars} & \rotatebox{90}{STL10} & \rotatebox{90}{SUN397} & {\cellcolor{tableblue}\textbf{Average}} \\
\midrule
TeCoA & 70.92 & 60.81 & 59.24 & 33.03 & 21.32 & 13.54 & 4.10 & 29.13 & 27.57 & 40.21 & 61.33 & 14.17 & 13.80 & 83.31 & 31.11 & \cellcolor{tableblue}37.57 \\
PMG-AFT & 77.48 & 62.21 & 64.84 & 37.45 & 24.41 & 14.71 & 5.00 & 30.92 & 38.29 & 33.44 & 64.79 & 18.77 & 19.18 & 86.31 & 30.43 & \cellcolor{tableblue}40.55 \\
TGA-ZSR & 71.60 & 63.82 & 62.90 & 35.85 & 20.26 & 12.86 & 4.16 & 26.45 & 27.90 & 52.39 & 64.98 & 30.61 & 14.67 & 84.92 & 30.56 & \cellcolor{tableblue}40.26 \\
\cellcolor{tableblue}\textbf{AGFT (Ours)} & \cellcolor{tableblue}{79.14} & \cellcolor{tableblue}{64.66} & \cellcolor{tableblue}{69.66} & \cellcolor{tableblue}{41.49} & \cellcolor{tableblue}{21.86} & \cellcolor{tableblue}{13.52} & \cellcolor{tableblue}{5.60} & \cellcolor{tableblue}{30.69} & \cellcolor{tableblue}{38.86} & \cellcolor{tableblue}{39.81} & \cellcolor{tableblue}{67.24} & \cellcolor{tableblue}{33.57} & \cellcolor{tableblue}{21.36} & \cellcolor{tableblue}{87.69} & \cellcolor{tableblue}{30.73} & \cellcolor{tableblue}\textbf{43.06} \\
\bottomrule
\end{tabular}%
}
\vspace{-6.5pt} 
\label{tab:12}
\end{table*}

\begin{table*}[t]
\captionsetup{justification=justified, singlelinecheck=false}
\caption{Zero-shot robust accuracy (\%).
Adversarial examples are generated by \textbf{AutoAttack} with
the perturbation budget $\epsilon = 1/255$.}
\vspace{-9pt} 
\centering
\resizebox{\linewidth}{!}{%
\renewcommand{\arraystretch}{0.9} 
\begin{tabular}{c c c c c c c c c c c c c c c c c}
\toprule
\textbf{Method} & \rotatebox{90}{Caltech101} & \rotatebox{90}{Caltech256} & \rotatebox{90}{CIFAR10} & \rotatebox{90}{CIFAR100} & \rotatebox{90}{DTD} & \rotatebox{90}{EuroSAT} & \rotatebox{90}{FGVC} & \rotatebox{90}{Flower102} & \rotatebox{90}{Food101} & \rotatebox{90}{ImageNet} & \rotatebox{90}{OxfordPet} & \rotatebox{90}{PCAM} & \rotatebox{90}{Stanf.Cars} & \rotatebox{90}{STL10} & \rotatebox{90}{SUN397} & {\cellcolor{tableblue}\textbf{Average}} \\
\midrule
TeCoA & 70.62 & 60.22 & 58.95 & 32.80 & 21.32 & 12.75 & 3.68 & 28.53 & 26.63 & 39.04 & 60.81 & 14.56 & 12.06 & 83.13 & 30.08 & \cellcolor{tableblue}37.01 \\
PMG-AFT & 77.06 & 61.89 & 64.75 & 37.07 & 24.41 & 14.11 & 4.58 & 30.61 & 37.16 & 32.83 & 64.41 & 19.22 & 17.42 & 86.17 & 29.67 & \cellcolor{tableblue}40.09 \\
TGA-ZSR & 71.37 & 63.46 & 62.62 & 35.49 & 20.31 & 12.68 & 3.74 & 26.16 & 27.19 & 51.56 & 64.79 & 31.04 & 13.41 & 84.83 & 29.79 & \cellcolor{tableblue}39.90 \\
\cellcolor{tableblue}\textbf{AGFT (Ours)} & \cellcolor{tableblue}{79.02} & \cellcolor{tableblue}{64.53} & \cellcolor{tableblue}{69.73} & \cellcolor{tableblue}{41.37} & \cellcolor{tableblue}{22.02} & \cellcolor{tableblue}{12.99} & \cellcolor{tableblue}{5.18} & \cellcolor{tableblue}{30.58} & \cellcolor{tableblue}{38.18} & \cellcolor{tableblue}{39.34} & \cellcolor{tableblue}{67.24} & \cellcolor{tableblue}{33.71} & \cellcolor{tableblue}{20.11} & \cellcolor{tableblue}{87.81} & \cellcolor{tableblue}{30.30} & \cellcolor{tableblue}\textbf{42.81} \\
\bottomrule
\end{tabular}%
}
\vspace{-6.5pt} 
\label{tab:13}
\end{table*}

\begin{table*}[t]
\captionsetup{justification=justified, singlelinecheck=false}
\caption{Zero-shot robust accuracy (\%) on \textbf{ViT-B/16}.
Adversarial examples are generated by PGD-20 attack with ${\epsilon = 1/255}$.}
\vspace{-9pt} 
\centering
\resizebox{\linewidth}{!}{%
\renewcommand{\arraystretch}{0.9} 
\begin{tabular}{c c c c c c c c c c c c c c c c c}
\toprule
\textbf{Method} & \rotatebox{90}{Caltech101} & \rotatebox{90}{Caltech256} & \rotatebox{90}{CIFAR10} & \rotatebox{90}{CIFAR100} & \rotatebox{90}{DTD} & \rotatebox{90}{EuroSAT} & \rotatebox{90}{FGVC} & \rotatebox{90}{Flower102} & \rotatebox{90}{Food101} & \rotatebox{90}{ImageNet} & \rotatebox{90}{OxfordPet} & \rotatebox{90}{PCAM} & \rotatebox{90}{Stanf.Cars} & \rotatebox{90}{STL10} & \rotatebox{90}{SUN397} & {\cellcolor{tableblue}\textbf{Average}} \\
\midrule
TeCoA & 76.71 & 66.03 & 60.51 & 35.74 & 25.37 & 15.57 & 5.48 & 36.27 & 30.62 & 48.86 & 66.15 & 19.43 & 19.72 & 88.03 & 36.20 & \cellcolor{tableblue}42.05 \\
PMG-AFT & 83.34 & 69.42 & 71.82 & 43.52 & 26.38 & 10.29 & 7.28 & 38.01 & 42.18 & 44.87 & 69.04 & 48.44 & 24.26 & 90.92 & 36.47 & \cellcolor{tableblue}47.08 \\
TGA-ZSR & 75.91 & 67.11 & 65.31 & 38.58 & 24.30 & 17.39 & 5.39 & 27.25 & 29.04 & 60.35 & 68.25 & 33.93 & 16.00 & 89.86 & 35.80 & \cellcolor{tableblue}43.63 \\
\cellcolor{tableblue}\textbf{AGFT (Ours)} & \cellcolor{tableblue}{85.60} & \cellcolor{tableblue}{71.75} & \cellcolor{tableblue}{73.40} & \cellcolor{tableblue}{48.39} & \cellcolor{tableblue}{25.10} & \cellcolor{tableblue}{8.25} & \cellcolor{tableblue}{9.77} & \cellcolor{tableblue}{42.61} & \cellcolor{tableblue}{45.55} & \cellcolor{tableblue}{51.56} & \cellcolor{tableblue}{73.67} & \cellcolor{tableblue}{49.44} & \cellcolor{tableblue}{31.98} & \cellcolor{tableblue}{91.41} & \cellcolor{tableblue}{39.12} & \cellcolor{tableblue}\textbf{49.84} \\
\bottomrule
\end{tabular}%
}
\vspace{-6.5pt} 
\label{tab:14}
\end{table*}

\begin{table*}[t]
\captionsetup{justification=justified, singlelinecheck=false}
\caption{Zero-shot robust accuracy (\%) on \textbf{RN50}$\bf{\times}$\textbf{4}.
Adversarial examples are generated by PGD-20 attack with ${\epsilon = 1/255}$.}
\vspace{-9pt} 
\centering
\resizebox{\linewidth}{!}{%
\renewcommand{\arraystretch}{0.9} 
\begin{tabular}{c c c c c c c c c c c c c c c c c}
\toprule
\textbf{Method} & \rotatebox{90}{Caltech101} & \rotatebox{90}{Caltech256} & \rotatebox{90}{CIFAR10} & \rotatebox{90}{CIFAR100} & \rotatebox{90}{DTD} & \rotatebox{90}{EuroSAT} & \rotatebox{90}{FGVC} & \rotatebox{90}{Flower102} & \rotatebox{90}{Food101} & \rotatebox{90}{ImageNet} & \rotatebox{90}{OxfordPet} & \rotatebox{90}{PCAM} & \rotatebox{90}{Stanf.Cars} & \rotatebox{90}{STL10} & \rotatebox{90}{SUN397} & {\cellcolor{tableblue}\textbf{Average}} \\
\midrule
TeCoA & 58.61 & 50.81 & 34.97 & 18.23 & 22.92 & 8.77 & 3.53 & 17.85 & 16.07 & 33.12 & 58.82 & 32.45 & 11.81 & 76.71 & 20.29 & \cellcolor{tableblue}31.00 \\
PMG-AFT & 63.58 & 52.15 & 31.02 & 10.70 & 20.37 & 10.67 & 4.88 & 21.00 & 24.02 & 31.73 & 59.28 & 14.03 & 16.84 & 68.02 & 25.66 & \cellcolor{tableblue}30.26 \\
TGA-ZSR & - & - & - & - & - & - & - & - & - & - & - & - & - & - & - & \cellcolor{tableblue}- \\
\cellcolor{tableblue}\textbf{AGFT (Ours)} & \cellcolor{tableblue}{74.09} & \cellcolor{tableblue}{59.43} & \cellcolor{tableblue}{35.31} & \cellcolor{tableblue}{14.69} & \cellcolor{tableblue}{22.44} & \cellcolor{tableblue}{9.84} & \cellcolor{tableblue}{5.15} & \cellcolor{tableblue}{25.36} & \cellcolor{tableblue}{36.72} & \cellcolor{tableblue}{39.00} & \cellcolor{tableblue}{64.93} & \cellcolor{tableblue}{16.72} & \cellcolor{tableblue}{22.37} & \cellcolor{tableblue}{76.07} & \cellcolor{tableblue}{28.19} & \cellcolor{tableblue}\textbf{34.69} \\
\bottomrule
\end{tabular}%
}
\vspace{-6.5pt} 
\label{tab:15}
\end{table*}

\end{document}